\documentclass{article}
\usepackage{ijcai22}

\usepackage{times}
\usepackage{soul}
\usepackage{paralist}
\usepackage{url}
\usepackage[hidelinks]{hyperref}
\usepackage[utf8]{inputenc}
\usepackage[small]{caption}
\usepackage{graphicx}
\usepackage{amsmath}
\usepackage{amsthm}
\usepackage{amssymb}
\usepackage{mathrsfs}
\usepackage{stmaryrd}
\usepackage{float}
\usepackage{color}
\usepackage{changes}
\usepackage{xspace}

\usepackage{booktabs}
\usepackage{multirow}
\usepackage{stfloats}
\usepackage{lipsum}

\usepackage{algorithm}
\usepackage{algorithmic}
\usepackage{subfiles}
\usepackage{amsfonts}

\usepackage{bbding}
\newcommand{\vic}[1]{\textcolor{black}{#1}}
\newcommand{\vicf}[1]{\textcolor{black}{#1}}
\newcommand{\zhiwei}[1]{\textcolor{black}{#1}}

\newcommand{\jeff}[1]{\textcolor{black}{#1}}

\title{
Type-aware Embeddings for Multi-Hop Reasoning over Knowledge Graphs}

 \author{
 Zhiwei Hu$^1$\and 
 Víctor Gutiérrez-Basulto$^{2}$\and
 Zhiliang Xiang$^{2}$\and
 Xiaoli Li$^3$ \and
 Ru Li$^1$\footnote{Contact Authors}\and
 Jeff Z. Pan$^{4*}$ 
 \\
 \affiliations
 $^1$School of Computer and Information Technology, Shanxi University, China\\
 $^2$School of Computer Science and Informatics, Cardiff University, UK\\
 $^3$Institute for Infocomm Research/Centre for Frontier AI Research, A*STAR, Singapore\\
 $^4$ILCC, School of Informatics, University of Edinburgh, UK\\
 \emails
 zhiweihu@whu.edu.cn,
 \{gutierrezbasultov, xiangz6\}@cardiff.ac.uk,
 xlli@i2r.a-star.edu.sg,
 liru@sxu.edu.cn, j.z.pan@ed.ac.uk
 }

\begin{document}

\maketitle
  
\begin{abstract}
Multi-hop reasoning over \vic{real-life} knowledge graphs (KGs)  is a \vic{highly} challenging \vic{problem as traditional subgraph matching methods are not capable to deal with noise and missing information.} 
\vicf{To address this problem, it has been recently  introduced a promising approach} based on \vic{jointly embedding logical queries and KGs} 
into a  low-dimensional space  to identify answer entities. 
However, \vic{existing proposals} ignore critical \vic{semantic} knowledge \vic{inherently available in KGs}, such as type information. 
\vicf{To leverage  type information, 
we propose a novel  \textbf{T}yp\textbf{E}-aware \textbf{M}essage \textbf{P}assing (\textsc{Temp}) model}, which enhances the entity and relation \jeff{representations} in queries, and simultaneously improves  generalization, deductive and inductive reasoning. 
\vic{Remarkably,} \textsc{Temp} is a \vic{plug-and-play} model that can be easily incorporated into  existing \vic{embedding-based} models to improve their performance. 
Extensive experiments on three real-world datasets 
demonstrate \vic{\textsc{Temp}'s} effectiveness. 
\end{abstract}

\section{Introduction}

\begin{figure}[t!]
    \centering
    \includegraphics[width=0.45\textwidth]{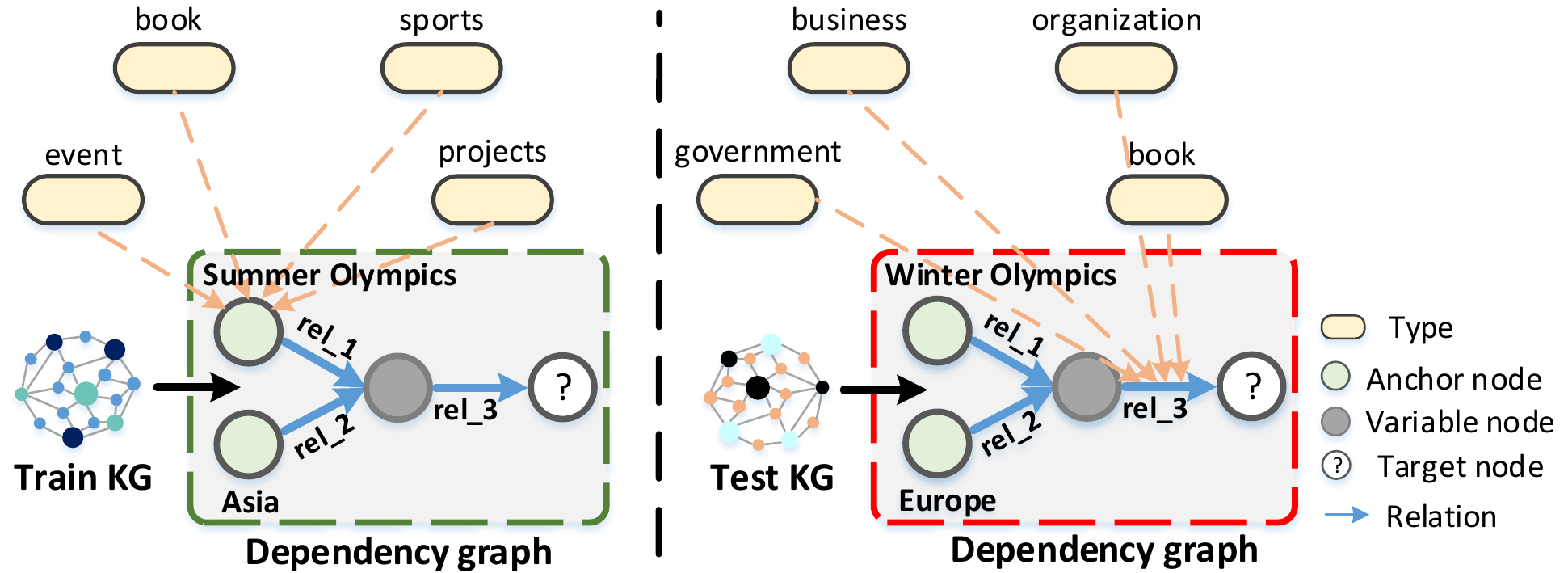}
    \caption{Inductive setting for FOL queries. The left part shows the  query ``\emph{List the presidents of Asian countries that have held the Summer Olympics}". The right part shows the query ``\emph{List the presidents of European countries that have held the Winter Olympics}". 
    \emph{rel\_1}, \emph{rel\_2}, and \emph{rel\_3} represent  relations: \emph{Hold}, \emph{Locate}, and \emph{President of}, respectively. } 
    \label{fig_1}
\end{figure}

\vic{In recent years, the multi-hop reasoning problem of answering first-order logic queries (FOL) on large-scale incomplete knowledge graphs (KGs)~\cite{Pan2016}  has  gained a lot of attention in the AI community.}
%
%
\vicf{A major challenge for traditional subgraph matching methods for query answering is that KGs  are inevitably incomplete and noisy.}
Indeed, when schema~\cite{WPKD2020} and   triples are incomplete in the KG, correct answers are not guaranteed to be found under normal deductive reasoning, leading to empty or wrong answers. Another problem is their intrinsic high computational complexity  as they need to  keep track of all intermediate entities occurring on reasoning paths, leading to an exponential blow-up. For instance, to answer the query \textit{``List the presidents of Asian countries that have held the Summer Olympics"} shown in Fig.\ \ref{fig_1}, we require two traversing-steps (many more for other queries): one to identify countries that have held the summer Olympics and another one to identify Asian countries, each producing intermediate countries.

 \vicf{To address these challenges, a \emph{query embedding (QE)} approach to query answering has been recently introduced as an alternative to subgraph matching methods. The main idea 
 is  to embed entities and  queries into a joint low-dimensional vector space such that entities that answer the query are close to the embedding of the query.} Several QE models for  query answering, showing very promising performance,  have been proposed so far~\cite{hamilton2018embedding,query2box:ren2020,ren2020beta,zhang2021cone,choudhary2021self,luus2021logic}. However, these models \textit{fail to leverage semantic knowledge inherently available in KGs}, such as entity description~\cite{yao2019kg,daza2021inductive} or entity type information~\cite{niu2020autoeter,pan2021context}. 
Advantages of introducing type information are that:~1) it can enhance the representation of entities or relations; e.g., 
the types \emph{sports}
and \emph{event} 
can enrich the representation of the entity \emph{Summer Olympics} in the context of sport events (cf.\  Fig.~\ref{fig_1}). 2) \zhiwei{I}t can also  help tackling the \textit{inductive query answering} problem where entities used in  test queries cannot be observed at training time; e.g.,  
consider the queries  in Figure \ref{fig_1}: ``\emph{List the presidents of Asian countries that have held the Summer Olympics}" and ``\emph{List the presidents of European countries that have held the Winter Olympics}", which are generated from two KGs with disjoint sets of entities: \emph{Train KG} and \emph{Test KG}, respectively. 
Even if  the entities \emph{Summer Olympics} and \emph{Winter Olympics}  are different, they have similar type information, such as \emph{sports} and \emph{event}. 
Consequently, after using type information to represent  entities, the model associated to the query generated from \emph{Train KG} is also effective on the  query generated from  \emph{Test KG}.
\par \vicf{The   goal of this paper is to introduce a \emph{type-aware} \emph{plug-and-play} model which  makes full use of type information in the knowledge graph, 
and can be easily embedded into existing QE-based models. 
To this aim, we propose a novel   \textbf{T}yp\textbf{E}-aware \textbf{M}essage \textbf{P}assing (\textsc{Temp}) model,} 
which contains two key components. 1) Type-aware Entity Representations (TER), aggregating  type information of entities to strengthen their \vicf{vector} representation (\vicf{cf. Section~\ref{sec:TER}}). 2) Type-aware Relation Representations (TRR), using entity type information to construct a global \textit{type graph} to enhance the relation representation, and simultaneously integrate it with its \vicf{type} representation and  existing entity type information (\vicf{cf. Section~\ref{sec:TRR}}).
\vicf{Importantly}, some queries have \textit{variable nodes} in the query paths (see Figure \ref{fig_1}), 
which increase the difficulty of subsequent  reasoning steps in the chain, as variable nodes are unknown. \vicf{To address this,  the TRR component uses a bidirectional mechanism} for  the \textit{anchor node to supervise the relations} in the query path, and vice versa. 
\vicf{Furthermore}, as mentioned,  after using type information to represent entities and relations, the model becomes inherently inductive as the occurrence of new entities or relations will not affect the type-based representations. 
\par Our main contributions can be summarized as follows:
\begin{compactitem}
\item \vic{We propose \textsc{Temp}, a novel \emph{type-aware plug-and-play} model for multi-hop reasoning \jeff{over}   KGs, 
that can be easily incorporated into the existing QE-based models.}
\item \vic{We design a new bidirectional integration mechanism that incorporates the pairwise dependencies among  \{entity, relation, type\} information, even in the absence of schema   axioms like domain and range.} 
\item \vic{Extensive experiments demonstrate that after incorporating \textsc{Temp} into four  state-of-the-art baselines, their \emph{ generalization}, \emph{deductive} and \emph{inductive reasoning}  abilities are significantly improved across three benchmark datasets consistently.}
\end{compactitem}

Data, code, and an extended version with appendix is available  at \url{https://github.com/zhiweihu1103/QE-TEMP}

\section{Related Work}
\paragraph{Query Embeddings.} QE models first embed entities and FOL queries into a joint low-dimensional vector space, and subsequently compute a similarity score between the \emph{entity representation} and \emph{query representation} in the latent embedding space. In general, according to the type of embedding spaces, QE-based methods can be divided into four categories: (i) geometric-based methods, \jeff{such as} GQE~\cite{hamilton2018embedding}, Q2B~\cite{query2box:ren2020}, HypE~\cite{choudhary2021self}, and ConE~\cite{zhang2021cone}, where logical queries and KG entities are embedded into a geometric vector space as points, boxes, hyperbolic, and cone shapes, respectively; (ii) distribution-based methods, such as \textsc{BetaE}~\cite{ren2020beta}, \jeff{embedding} queries to beta distributions with bounded support, and PERM~\cite{choudhary2021probabilistic}, \jeff{using} Gaussian densities to reason over KGs; 
(iii) logic-based methods, \vicf{relating so-called set logic with FOL}
~\cite{luus2021logic}; (iv) neural-based methods, \jeff{e.g.,} EMQL~\cite{sun2020faithful} \jeff{using} neural retrieval to implement logical operations. 
Considering QE-based methods \jeff{are} the mainstream  in the current CQA field, \textit{we mainly focus on how to construct a plug-and-play model to embed the type information for existing QE-based methods}.  

\paragraph{Other Methods.} Besides the QE-based approach, the path-based approach is another \vicf{method for} CQA, but it faces an exponential growth of the search \vicf{space} with the number of hops. For instance, CQD~\cite{arakelyan2020complex} uses a beam search method to explicitly track intermediate entities, and repeatedly combines scores from a pretrained link predictor via t-norms  to search answers while tracking intermediaries. However, CQD does not support the full set of FOL queries. 

\paragraph{Inductive KG Completion (KGC).} In the context of KGC, there have been some works on inductive settings where test entities are not seen in the training stage. Based on the source of information used, they can be split into two categories: Using  graph structure information,  e.g.,\ subgraph or topology structures~\cite{teru2020inductive,Jiajuninductive,wang2021relational}, or  using  external information, e.g.,\ textual descriptions of entities~\cite{daza2021inductive}. However, \textit{all these methods  focus on the inductive KGC task, which can be seen as answering simpler one-hop queries.}

\paragraph{Type-aware Tasks.} Type information was previously used in other tasks such as KGC or entity typing~\cite{yao2019kg,zhao2020connecting,daza2021inductive,niu2020autoeter,pan2021context}.
However, these works cannot be directly used for answering FOL queries because this  requires multi-hop reasoning, producing intermediate uncertain entities.

\section{Background}
In this paper, a knowledge graph~\cite{Pan2016} is represented in a standard format for graph-structured data such as RDF. A   \vicf{\emph{knowledge graph}} $\mathcal{G}$  is a tuple $(\mathcal{E},\mathcal{R}, \mathcal{C},\mathcal{T} )$, where  $\mathcal{E}$ is a set of entities,  $\mathcal{R}$ is a set of relation types, $\mathcal{C}$ is a set of entity types,  and  $\mathcal{T}$ is a set of triples.
Triples in $\mathcal{T}$ are 
either relation assertions $(h,r,t)$,
where $h,t \in \mathcal{E}$  are respectively the \emph{head} and \emph{tail} entities of the triple, and $r \in \mathcal{R}$ is the \emph{edge}  of the triple connecting  head and tail, or entity type assertions ($e$, $type$, $c$), where $e \in \mathcal{E}$ is an entity, $c\in \mathcal{C}$ is an entity type  and $type$ is  the instance-of relation~\cite{DBLP:series/ihis/Pan09}. 

\vic{We consider FOL queries that  use  existential quantification ($\exists$), conjunction ($\wedge$), disjunction ($\vee$) and negation ($\neg$) operations. We will work with FOL queries in Disjunctive Normal Form, i.e. represented as a disjunction of conjunctions. To introduce FOL queries, we assume that $\mathcal V_a \subset \mathcal E$ represents a set of  non-variable \emph{input anchor entities}, $V_1, \ldots, V_m$ denote \emph{existentially quantified variables} 
and $V_?$ is the \emph{target variable}. 
A FOL query $\mathcal Q$  is a formula of the following form: }
%
\begin{align*}
\mathcal{Q}[V_{?}] = & \,\, V_{?}\, . \, \exists V_{1}, \ldots, V_{m} : c_{1} \vee c_{2} \vee \ldots  \vee c_{n} 
\end{align*}
 %
\vic{ where  $c_{i} = e_{i1} \wedge \ldots \wedge$ $e_{ik}$, \emph{k} $\leq$ \emph{m} such that each $e_{ij}$ is of one of the following forms: $r(V_{a}, V)$, $\neg r(V_{a}, V)$,   $r(V_, V')$ or $\neg r(V, V')$, with $V_a \in \mathcal V_a$, $V \in \{V_{?}, V_{1},\ldots , V_{m}\}$, $V' \in \{V_{1},\ldots , V_{m}\}$,
   $V \neq V'.$}

 %
\vic{The \emph{dependency graph (DG) of a query $\mathcal Q$} is a graphical representation of  $\mathcal Q$, where nodes correspond to variable or non-variable arguments in $\mathcal Q$ and edges correspond to relations in $\mathcal Q$.  Figure~\ref{fig_1} shows an example of a DG.}

\vic{We are interested in the multi-hop reasoning problem of answering  queries $\mathcal Q$ on KGs, which aims to find a set of entities $\llbracket \mathcal{Q}\rrbracket \subseteq \mathcal E$ such that\ $a \in \llbracket \mathcal{Q}\rrbracket $ iff $\mathcal Q[a]$ holds true.} 

\section{Semantically-enriched Embeddings}
\begin{figure*}[!htp]
    \centering
    \includegraphics[width=0.95\textwidth]{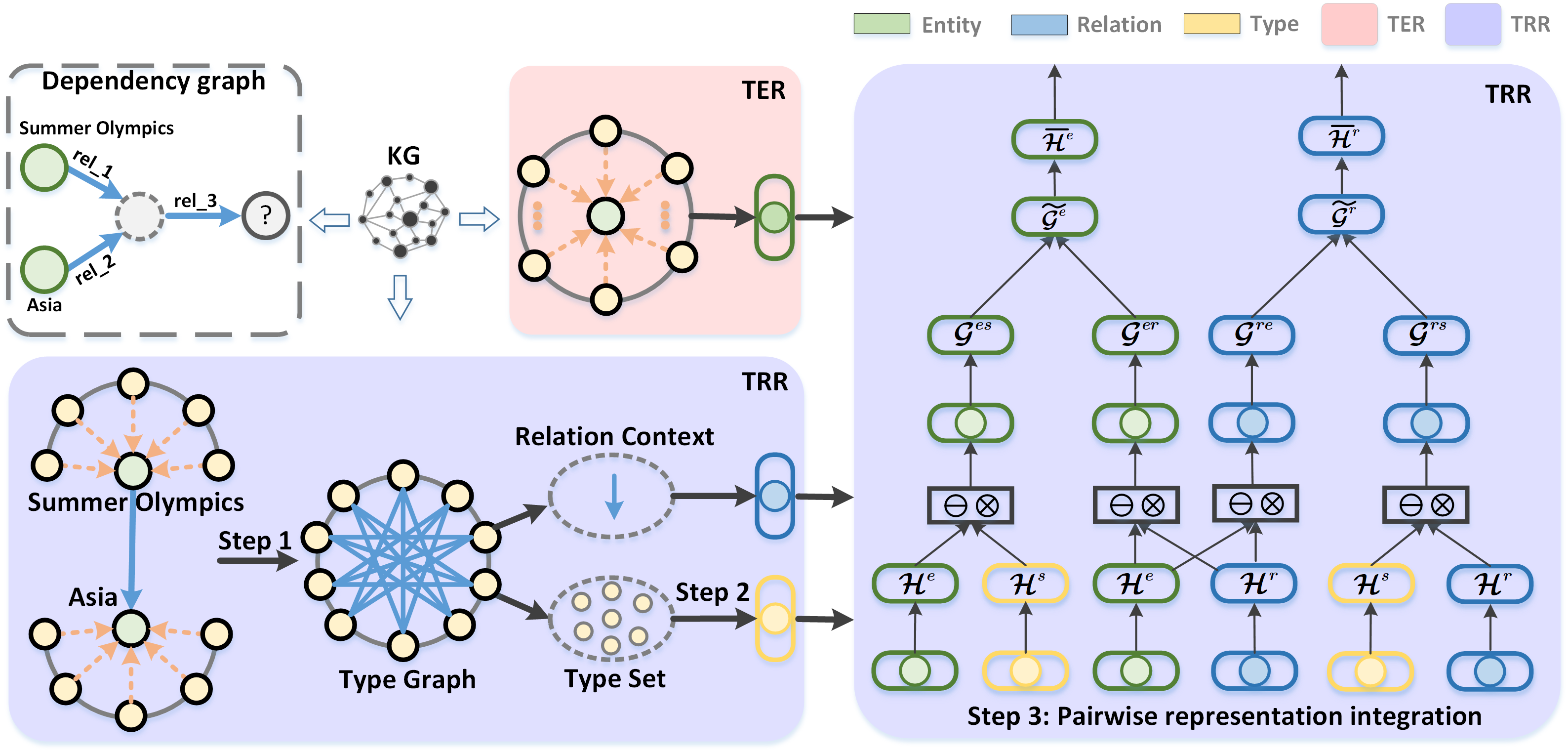}
    \caption{\textsc{Temp}'s Architecture. The left top  part is the dependency graph for query ``\emph{List the presidents of Asian countries that have held the Summer Olympics}", \emph{rel\_1}, \emph{rel\_2}, and \emph{rel\_3} represent the relations \emph{Hold}, \emph{Locate}, and \emph{President of}, respectively.} 
    \label{fig_3}
\end{figure*}
\vic{
\vicf{Our model} \textsc{Temp} is composed of  two sub-models:
\textit{Type-aware Entity Representations (TER)}, which uses  type information of an entity to enrich its vector representation, and \textit{Type-aware Relation Representations (TRR)}, which 
further integrates entity representations, relation types, and relation representations 
to strengthen the entity and relation vector representations simultaneously. 
Interestingly, as we only leverage type information to perform an in-depth characterization of entities and relations \emph{without modifying the training target} of existing QE-based models, \textsc{Temp} can be easily embedded into them in a plug-and-play fashion.} 
\subsection{TER: Type-aware Entity Representations}\label{sec:TER}
\vic{ The main intuition behind TER is that the types of an entity provide 
valuable information about what it represents in the KG. 
For instance, if an entity contains types such as \textit{sports/multi\_event\_tournament, time/event, olympics/olympic\_games}, it is  plausible to infer that the corresponding entity represents 
\textit{Olympics}. To capture this intuition, we design an iterative  multi-highway layer~\cite{highway:2015} 
to aggregate the type information in entity type assertions 
to get a more accurate and comprehensive representation of it\footnote{See appendix for other aggregation alternatives.}. Let  $\mathcal{H}_{s}^{i}\in\mathbb{R}^{d\times n}$ denote the hidden state of type information of an entity 
in iteration $i \geq 1$, where \emph{d} and \emph{n} respectively represent the vector size and the number of types of an entity. The highway-based type fusion representation of a given entity can be calculated as follows:}

\vspace{-0.2cm}
\begin{equation}
g = \sigma(W_{i}\mathcal{H}_{s}^{i} + b_{i})
\label{highway_agg_1}
\end{equation}
\vspace{-0.3cm}
\begin{equation}
\mathcal{H}_{s}^{i+1} = g\ast(W_{i}'\mathcal{H}_{s}^{i} + b_{i}') + (1-g)\ast\mathcal{H}_{s}^{i}
\label{highway_agg_2}
\end{equation}
\vspace{-0.2cm}
\begin{equation}
\widetilde{\mathcal{H}_{s}^{K}} = W\mathcal{H}_{s}^{K} + b
\label{highway_agg_3}
\end{equation}
\zhiwei{ $\mathcal{H}_{s}^{1}$} is the initial feature of types of  an entity, $\sigma$ is an element-wise sigmoid function, \{$W_{i}, W_{i}'$\}$\in\mathbb{R}^{d\times d}$, \{$b_{i}, b_{i}'$\}$\in\mathbb{R}^{d\times 1}$ are  learnable matrices, and \emph{g}$\in\mathbb{R}^{d\times n}$ is the reset gate. After iterating \emph{K} times \zhiwei{(we set \emph{K}=2)}, the final message $\mathcal{H}_{s}^{K}\in\mathbb{R}^{d\times n}$ (undergoing a linear operation to obtain $\widetilde{\mathcal{H}_{s}^{K}}\in\mathbb{R}^{d\times 1}$) is taken as the representation for the  types of a given entity. 
%
 \vic{We further concatenate the initial entity and its type aggregation representation to get an enhanced entity representation.}
\begin{equation} 
\mathcal{H}^{e} = W'[\widetilde{\mathcal{H}_{s}^{K}}, \mathcal{H}_{s}^{0}] + b',
\label{resnet_entity}
\end{equation}   
 \vic{where [$\cdot$] is the concatenation function, $W^{'}$$\in\mathbb{R}^{d\times d}$ and $b^{'}$$\in\mathbb{R}^{d\times 1}$ are the parameters to learn. $\mathcal{H}^{e}\in\mathbb{R}^{d\times 1}$ is the final representation of the entity. It is important to note that for inductive reasoning, we will not concatenate with the initial entity information $\mathcal{H}_{s}^{0}$ as the entities seen during training are not presented in the test phase. The process is shown in top center of Figure~\ref{fig_3}.}
\vspace{-0.2cm}


\subsection{TRR: Type-aware Relation Representations}\label{sec:TRR}
\vic{Performing TER on entities is useful for queries without existentially quantified variables. However, for queries with \emph{chained}  existential variables (chain of variable nodes in the DG) it is not enough to only perform TER on the anchor entity \zhiwei{or} target variable. Intuitively, the problem is that in the long-chain reasoning process, the  correlation between the anchor entity  and target variable may not be strong enough after several relation projections. Besides, continuous relational projections may cause exponential growth in the search space, further increasing the complexity of \zhiwei{the model}.}
\par 

\vic{We start by observing that the types of a relation  are correlated with its representation in the KG. For example, assuming a relation $r$ has types  \textit{government/ political\_appointer} and \textit{organization/role}, then we can plausibly infer that the relation  $r$ represents  \textit{President of}.
In long-chain queries, type-enhancement on relations can help to reduce the answer entity space and cascading errors caused by multiple projections. However,  in most existing KGs,  relations lack  specific type annotations (such as domain and range). We address this problem 
by building, based on the original KG, a novel 
\emph{type graph} with types as nodes and relations as edges (see bottom left of Fig.~\ref{fig_3}). 
In a subsequent step, we aggregate the type information on the type graph to obtain the type embedding corresponding to a specific relation. Finally, we integrate the entity representation, the aggregated type information of a  relation and its 
representation by a bidirectional attention mechanism, so that the intermediate variable nodes can perceive the message of anchor or target nodes and of the relations in the chain of reasoning (see right of Fig.~\ref{fig_3}). This will help to avoid the weakening of the connection between anchor and target entity caused by long-chain reasoning.}
\subsubsection{Step 1: Type Graph Construction}
\vic{We formally define a \emph{type graph $\mathcal G_\mathsf{tp}$}. }
 Let  $\mathcal G = (\mathcal{E,R,C,T})$ be a KG. For a  relation $r\in \mathcal{R}$, 
 $\mathcal T_r\subseteq \mathcal T$  denotes the set of  relation assertions in which $r$ occurs. For a  relation assertion $\mathsf t\in  \bigcup_{r \in \mathcal R}\mathcal T_r$, $\mathsf{hd}(\mathsf t)$ and $\mathsf{tl}( \mathsf t)$   respectively denote 
the head  and tail entities of $\mathsf t$, and $\mathsf{tp}_\mathsf t (\mathsf{hd}(\mathsf t)) = \{c \mid (\mathsf{hd}(\mathsf t), type, c) \in \mathcal T\}$ denotes the set of types of  the head   of $\mathsf t$; $\mathsf{tp}_{\mathsf t}(\mathsf{tl}( \mathsf t))$ is defined analogously. 
Since  $r$ may occur in multiple relation assertions, we will compute the type information of $r$ by taking the intersection of the types of the head and tail entities of relation assertions in which  $r$ occurs. 
For $r \in \mathcal R$, 
we define
\begin{equation*}
    \mathsf{tp}_{r}^{\mathsf{hd}}(\mathcal G) = \bigcap_{ \mathsf t \in \mathcal T_r} \mathsf{tp}_\mathsf t(\mathsf{hd}(\mathsf t)), \qquad \mathsf{tp}_{r}^{\mathsf{tl}}(\mathcal G) = \bigcap_{\mathsf t \in \mathcal T_r} \mathsf{tp}_\mathsf t(\mathsf{tl (t)}).\vspace{-0.1cm}
     \end{equation*}
\vic{
In addition, we define $\mathcal G_\mathsf{tp}=(V, E, T)$ by setting 
$V = \bigcup_{r \in \mathcal R} \mathsf{tp}^{\mathsf{hd}}_{r}(\mathcal G) \cup \mathsf{tp}^{\mathsf{tl}}_{r}(\mathcal G)$ 
\zhiwei{,} $E=\mathcal R$, and $(v,r,v') \in T$ if there exists $\mathsf t \in \mathcal T_r$ such that $v = \mathsf{tp}_\mathsf t(\mathsf{hd(t)})$ and $v' = \mathsf{tp}_\mathsf t(\mathsf{tl(t)})$.}  
\subsubsection{Step 2: Relation Type Aggregation}
\vic{For a given  relation $r \in E$, 
we define the types associated with a relation as $\mathsf{tp}_r(\mathcal G_\mathsf{tp})$ =  $\mathsf{tp}^{\mathsf{hd}}_{r}(\mathcal G) \cup \mathsf{tp}^{\mathsf{tl}}_{r}(\mathcal G)$. We fix an arbitrary linear order on the elements of $\mathsf{tp}_r(\mathcal G_\mathsf{tp})$, 
and denote by $\mathsf{tp}^i_r(\mathcal G_\mathsf{tp})$ the $i$-th type, for all $1 \leq i \leq |\mathsf{tp}_r(\mathcal G_\mathsf{tp})|$.  Note that not all types in $\mathsf{tp}_r(\mathcal G_\mathsf{tp})$ are relevant for answering a given query. For example, assume that  the relation \emph{has\_part} contains the types \{\emph{vehicle}, \emph{animal}, \emph{universe}\}. For the query \emph{``What organs are parts of a cat?"}, we should give type \emph{animal} more attention, but for the query \emph{``What components are parts of a car?"} we should concentrate on the type \emph{vehicle}.}
\vic{So, instead of simply concatenating (or averaging) all the type information associated to a relation, we model the relation type aggregation as an attention neural network, 
defined as: \\}
\vspace{-0.2cm}
\begin{equation}
\mathcal{H}^{s} = \sum_{i} a_{i} \odot \mathcal{H}_{s}^{i}
\label{type_aggr}
\end{equation}
\vspace{-0.1cm}
\begin{equation}
a_{i} = \frac{exp(\textbf{MLP}(\mathcal{H}_{r}^{i}))}{\sum_{j}exp(\textbf{MLP}(\mathcal{H}_{r}^{j}))}
\end{equation}
\vic{ $\mathcal{H}^{s}$ is the vector representation of the aggregated type information $\mathsf{tp}_r(\mathcal G_\mathsf{tp})$; $\mathcal{H}_{s}^{i}\in{\mathbb{R}^{d\times1}}$ is the  vector representation of the \emph{i-th} type $\mathsf{tp}_r^i(\mathcal G_\mathsf{tp})$, which is initialized to a uniform distribution with dimension \emph{d}, $1 \leq i \leq |\mathsf{tp}_r^i(\mathcal G_\mathsf{tp})|$; $a_{i}\in{\mathbb{R}^{d\times1}}$ is a positive weight vector that satisfies $\sum_{i=1}^{n}[a_i]_{j}=1$ for all $ 1 \leq j \leq d$; 
and \textbf{MLP}($\cdot$) : $\mathbb{R}^{d} \rightarrow \mathbb{R}^{d}$ is a multi-layer perceptron network.}
\subsubsection{Step 3: Pairwise Representation Integration}
\vic{When embedding  queries, integrating the information of entities, relations, and types can help to smooth decision boundaries, but  
this needs to be done in a way that the intended match of the query into the KG is captured. For example, for  the query  \emph{``Which countries have held the Summer Olympics}?", 
we need to concentrate on \emph{Held} connections from  \emph{Summer Olympics}, rather than e.g.,\ \emph{Watch} connections. Analogously, we should only consider \emph{Held} connections starting at  \emph{Summer Olympics}, rather than e.g.,\ at \emph{World Cup}.}
\vic{To  properly capture this restriction 
in the triple \{$\mathcal{H}^{e}$, $\mathcal{H}^{r}$, $\mathcal{H}^{s}$\}  ($\mathcal{H}^{e}$ and $\mathcal{H}^{s}$ defined as in Equations~\eqref{resnet_entity} and \eqref{type_aggr}, and $\mathcal{H}^{r}$ is the initialization relation vector), 
we introduce a \textit{bidirectional attention mechanism}~\cite{zhang2020dcmn} to integrate each state of pairwise representation pairs: \emph{entity-relation, entity-type,} and \emph{relation-type}.  
Here, we show how to do this for entity-relation \zhiwei{pair}.
 Bidirectional integration representation between 
 $\mathcal{H}^{e}$ and 
 $\mathcal{H}^{r}$ can be calculated as follows:}
\begin{equation}
\mathcal{G}^{er} = Relu(W_{1}\begin{bmatrix}
\mathcal{H}^{e} \ominus \mathcal{H}^{r}\\
\mathcal{H}^{e} \otimes \mathcal{H}^{r}
\end{bmatrix} + b_{1})
\end{equation}
\vspace{-0.1cm}
\begin{equation}
\mathcal{G}^{re} = Relu(W_{2}\begin{bmatrix}
\mathcal{H}^{r} \ominus \mathcal{H}^{e}\\
\mathcal{H}^{r} \otimes \mathcal{H}^{e}
\end{bmatrix} + b_{2})
\end{equation}
\vic{ $\{W_{1}, W_{2}\} \in \mathbb{R}^{2d\times 2d}$ and \{$b_{1}, b_{2}\} \in \mathbb{R}^{2d\times 1}$ are learnable parameters. 
We use element-wise subtraction $\ominus$ and multiplication $\otimes$ to build better matching representations~\cite{tai2015improved}. $\mathcal{G}^{er} \in \mathbb{R}^{2d\times 1}$ is the result of integrating entity relation information. Through bidirectional integration of entities and relations, we simultaneously  get a relation-aware entity representation and an entity-aware relation representation, capturing the interaction between entities and relations.}

 \vic{We  then use a gated mechanism to combine the results produced by bidirectional fusion as it better regulates the information flow~\cite{highway:2015}. Take the entity fusion representation as an example, using the  relation-aware entity $\mathcal{G}^{er}$ and type-aware entity $\mathcal{G}^{es}$ representations as input, the final representation of entity is computed as }
\vspace{-0.1cm}
\begin{equation}
g = \sigma(W_{3}\mathcal{G}^{er} + W_{4}\mathcal{G}^{es} + b_{3} + b_{4})
\label{dcmn_1}
\end{equation}
\vspace{-0.3cm}
\begin{equation}
\widetilde{\mathcal{G}^{e}} = g\ast\mathcal{G}^{er} + (1 - g)\ast\mathcal{G}^{es}
\label{dcmn_2}
\end{equation}
 \vic{\{$W_{3}$, $W_{4}$\} $\in$ $\mathbb{R}^{2d\times2d}$ and \{$b_{3}$, $b_{4}$\} $\in$ $\mathbb{R}^{2d\times1}$ are the parameters to learn. \emph{g} is the reset gate, and  $\widetilde{\mathcal{G}^{e}}\in \mathbb{R}^{2d\times1}$ is the final entity representation.} 

\vic{To transform the fused feature to the original vector size, we use one linear layer: $\overline{\mathcal{H}^{e}} = W_{5}\widetilde{\mathcal{G}^{e}} + b_{5}$,
 where $W_{5}$ $\in$ $\mathbb{R}^{d\times2d}$ and $b_{5}$ $\in$ $\mathbb{R}^{d\times1}$ are learnable parameters. $\overline{\mathcal{H}^{e}}$ is the final  entity representation enhanced by relation and type.}

\section{Experiments}
\begin{table*}[!htp]
\renewcommand\arraystretch{0.50}
\setlength{\tabcolsep}{0.95em}
\centering
\begin{tabular*}{\linewidth}{@{}ccccccccccccc@{}}
\toprule
\multicolumn{11}{c|}{(a) \textbf{Generalization on FB15k-237} (Q2B datasets)} & \textbf{FB15k} & \textbf{NELL}\\
\midrule
\multicolumn{1}{c|}{\textbf{Method}} & \textbf{1p} & \textbf{2p} & \textbf{3p} & \textbf{2i} & \multicolumn{1}{c|}{\textbf{3i}} & \textbf{pi} & \textbf{ip} & \textbf{2u} & \textbf{up} & \multicolumn{1}{c|}{\textbf{Avg}} & \textbf{Avg} & \textbf{Avg} \\
\midrule
\multicolumn{1}{l}{GQE}     
& 41.3      & 21.5      & 15.2      & 26.5      & 38.5      & 16.7      & 8.8      & 17.1   & 15.8      & 22.4      & 40.1      & 23.5 \\
\multicolumn{1}{c}{+\textsc{Temp}}     
& \textbf{47.6}      & \textbf{29.6}      & \textbf{24.7}      & \textbf{36.3}        & \textbf{48.4}      & \textbf{25.5}      & \textbf{13.4}        & \textbf{30.2}      & \textbf{21.0}      & \textbf{30.7}        & \textbf{56.6}      & \textbf{38.2} \\
\midrule
\multicolumn{1}{l}{Q2B}     
& \textbf{47.1}      & 24.9      & 19.4      & 33.2      & 46.4      & 21.8      & 11.3      & 25.3  & \textbf{19.3}      & 27.6      & 51.2      & 31.1 \\
\multicolumn{1}{c}{+\textsc{Temp}}     
& 45.7      & \textbf{27.8}      & \textbf{23.4}      & \textbf{36.9}     & \textbf{49.6}      & \textbf{22.9}      & \textbf{11.7}      & \textbf{27.6}  & 18.9      & \textbf{29.4}      & \textbf{55.4}      & \textbf{37.3} \\
\midrule
\multicolumn{1}{l}{\textsc{BetaE}}     
& 42.6      & 25.4      & 21.6      & 30.2      & 43.3      & 20.7      & 9.2      & 24.2  
& \textbf{18.3}      & 26.2      & 50.6      & 33.4 \\
\multicolumn{1}{c}{+\textsc{Temp}}     
& \textbf{43.3}      & \textbf{27.2}      & \textbf{22.7}      & \textbf{35.3} & \textbf{47.5}      & \textbf{24.3}      & \textbf{10.6}      & \textbf{26.7} & 18.2     & \textbf{28.4}      & \textbf{53.6}      & \textbf{34.5} \\
\midrule
\multicolumn{1}{l}{\textsc{LogicE}}     
& 45.6      & 27.8      & 24.1      & 34.7      & 46.5      & 23.5      & 12.0      & 27.1 
& 20.8      & 29.1      & 54.2      & \textbf{39.1} \\
\multicolumn{1}{c}{+\textsc{Temp}}     
& \textbf{46.6}      & \textbf{29.2}      & \textbf{25.0}      & \textbf{35.8} & \textbf{47.9}      & \textbf{24.9}      & \textbf{13.5}      & \textbf{28.7}
& \textbf{21.0}      & \textbf{30.3}      & \textbf{55.7}      & \textbf{39.1} \\
\midrule
\multicolumn{11}{c|}{(b) \textbf{Deductive Reasoning on FB15k-237} (Q2B datasets)} & \textbf{FB15k} & \textbf{NELL}\\
\midrule
\multicolumn{1}{l}{GQE}     
& 56.4      & 30.1      & 24.5      & 35.9      & 51.2      & 25.1      & 13.0      & 25.8 & 22.0      & 31.6      & 43.7      & 49.8 \\
\multicolumn{1}{c}{+\textsc{Temp}}     
& \textbf{76.3}      & \textbf{48.6}      & \textbf{39.0}      & \textbf{49.7}      
& \textbf{60.4}      & \textbf{36.9}      & \textbf{22.1}      & \textbf{59.0}
& \textbf{36.3}      & \textbf{47.6}      & \textbf{71.4}      & \textbf{75.5} \\
\midrule
\multicolumn{1}{l}{Q2B}     
& 58.5      & 34.3      & 28.1      & 44.7      & 62.1      & 23.9      & 11.7      & 40.5 & 22.0      & 36.2      & 43.7      & 51.1 \\
\multicolumn{1}{c}{+\textsc{Temp}}     
& \textbf{87.2}      & \textbf{59.6}      & \textbf{47.9}      & \textbf{67.2}      
& \textbf{72.7}      & \textbf{49.1}      & \textbf{29.8}      & \textbf{78.6} 
& \textbf{43.4}      & \textbf{59.5}      & \textbf{69.6}      & \textbf{90.4} \\
\midrule
\multicolumn{1}{l}{\textsc{BetaE}}     
& 77.9      & 52.6      & 44.5      & 59.0      & 67.8      & 42.2      & 23.5      & 63.7 
& 35.1      & 51.8      & \textbf{60.6}      & 80.2 \\
\multicolumn{1}{c}{+\textsc{Temp}}     
& \textbf{84.7}      & \textbf{58.3}      & \textbf{49.4}      & \textbf{62.3}      
& \textbf{68.8}      & \textbf{45.3}      & \textbf{28.5}      & \textbf{74.5}
& \textbf{41.1}      & \textbf{57.0}      & \textbf{60.6}      & \textbf{81.5} \\
\midrule
\multicolumn{1}{l}{\textsc{LogicE}}     
& 81.5      & 54.2      & 46.0      & 58.1      & 67.1      & 44.0      & 28.5      & 66.6 
& 40.8      & 54.1      & 65.5      & \textbf{85.3} \\
\multicolumn{1}{c}{+\textsc{Temp}}     
& \textbf{84.5}      & \textbf{59.8}      & \textbf{51.9}      & \textbf{59.3}      
& \textbf{68.1}      & \textbf{47.0}      & \textbf{33.4}      & \textbf{70.8}
& \textbf{45.4}      & \textbf{57.8}      & \textbf{67.1}      & 84.6 \\
\bottomrule
\end{tabular*}
\caption{Hits@3 results on the Q2B datasets testing generalization and deductive reasoning. Please see appendix  for full results on FB15k and NELL.}
\label{table_1}
\end{table*}

\vic{Our aim is to analyse the impact of adding \textsc{Temp} to existing QE models on their 
generalization, deductive and inductive reasoning abilities.}



\paragraph{Datasets and Queries.} 
For generalization and deductive reasoning, we use three standard benchmark KGs with official training/validation/test splits: FB15k~\cite{bordes2013translating}, FB15k-237~\cite{toutanova2015observed}, and NELL995 (NELL)~\cite{xiong2017deeppath}, and two query datasets: one with 9  query structures without negation from Query2Box (\textsc{Q2B})~\cite{query2box:ren2020} and another with 14  (9 positive + 5 with negation) from \textsc{BetaE}~\cite{ren2020beta}. To test inductive reasoning, we use the  datasets FB15k-237-V2 and NELL-V3 provided by GraIL~\cite{teru2020inductive},  ensuring that the test and training sets have a disjoint set of entities. Note that we generate queries over these datasets as done for \textsc{BetaE} datasets. We choose Hit@K and Mean Reciprocal Rank (MRR) as two evaluation metrics\footnote{Refer to appendix for  details on datasets, queries, and  metrics.}.
%


\paragraph{Baselines.} We embed  \textsc{Temp} on  four state-of-the-art baselines \vicf{for complex QA on KGs}: Q2B, \textsc{BetaE}, GQE~\cite{hamilton2018embedding}, and \textsc{LogicE}~\cite{luus2021logic}.

\paragraph{Generalization.} \vic{The goal is to find   \emph{non-trivial} answers to FOL queries over incomplete  KGs, i.e.\ answers cannot get using subgraph matching. We follow the evaluation protocol in~\cite{ren2020beta}. Briefly, three KGs are built: $\mathcal{G}_{\textit{train}}$  containing only training triples, $\mathcal{G}_{\textit{valid}}$ containing $\mathcal{G}_{\textit{train}}$ plus  validation triples, and $\mathcal{G}_{\textit{test}}$ containing $\mathcal{G}_{\textit{valid}}$ as well as  test triples. The models are trained using $\mathcal{G}_{\textit{train}}$ to evaluate the generalization ability because queries have at least one edge to predict to find an answer.}

\paragraph{Deductive Reasoning.} 
\vic{The goal is  to test the ability of finding  answers only using standard reasoning.
\vicf{Following~\cite{sun2020faithful}}, we train models using the full KG ($\mathcal{G}_{\textit{train}}\cup\mathcal{G}_{\textit{valid}}\cup\mathcal{G}_{\textit{test}}$), so only inference (not generalization) is used to get correct answers.} 
\paragraph{Inductive Reasoning.} \vic {All  baseline models have  inductive ability at the query  level as they can answer queries with structures that are not seen during training. For example, the  Q2B  and \textsc{BetaE} datasets consider five query structures during the training and validation phase and four `unseen' structures are used during testing.} 
\vicf{However, it is not known whether they have  \emph{entity-level} inductive ability, i.e.\  during testing, the query structure has \emph{entities that do not appear} in the training phase.} \vicf{We will analyse  this  for the first time.}

\subsection{Main Results}
\label{main_results}
\vic{We compare the performance of the four baseline models  with their counterparts after adding our \textsc{Temp} model in four different  aspects: 1) generalization, 2) deductive reasoning, 3) inductive reasoning, and 4) queries with negation.} 

\paragraph{Generalization.} \vic{Table \ref{table_1}(a) shows that for long-chain queries \emph{2p} and \emph{3p}, the improvement brought by \textsc{Temp} exceeds that of short-chain \emph{1p}, confirming the suitability of  type information for dealing with long-chain \vicf{queries}. In addition, \textsc{Temp}-enhanced models also achieve improvements on queries \emph{ip}/\emph{pi}/\emph{2u}/\emph{up}, which do not occur in the training KG, showing that type information is also helpful to improve \vicf{inductive ability at the query-level structure.}
Notably, for GQE,
adding type information can shorten the gap or even surpass the state-of-the-art baseline models (without \textsc{Temp}).}

\paragraph{Deductive Reasoning.} \vic{Table \ref{table_1}(b) shows that after adding  type information, the reasoning ability of the baselines are significantly improved on all datasets consistently. Specifically, the improvement of   embedding models based on geometric operations (GQE, Q2B) is more significant  than that of \textsc{BetaE} or \textsc{LogicE}.   
\vicf{Remarkably, Q2B {+} \textsc{Temp}  surpasses the state-of-the-art baseline models (without \textsc{Temp}).}
The main reason  \vicf{for the modest gain for \textsc{BetaE} and \textsc{LogicE} is that they} impose excessive restrictions on the embedding of entities and relations. For instance, in  \textsc{LogicE}, 
the logic embeddings with bounded support change the type-enriched vector representations, thus affecting the effect of type information.}

\begin{table}[!htp]
\setlength{\tabcolsep}{0.17em}
\centering
\small
\begin{tabular*}{\linewidth}{@{}cccccccccccccc@{}}
\toprule
\textbf{Data}& \textbf{Method} & \textbf{1p} & \textbf{2p} & \textbf{3p} & \textbf{2i} & \multicolumn{1}{c|}{\textbf{3i}} & \textbf{pi} & \textbf{ip} & \textbf{2u} & \textbf{up} & \textbf{Avg} \\
\midrule
\multirow{9}*{\rotatebox{90}{FB15k-237-V2}}
& \multicolumn{1}{l}{GQE}    
& 0.5   & 5.5    & 1.8    & 0.5    & 0.6 	& 7.1 	& 13.1 	& 2.1 	& 6.1 	& 4.1 \\
& \multicolumn{1}{c}{+\textsc{Temp}}  
& \textbf{14.6} 	& \textbf{22.1} 	& \textbf{14.1} 	& \textbf{13.9} 	
& \textbf{14.4} 	& \textbf{15.7}  	& \textbf{22.0} 	& \textbf{9.7}
& \textbf{20.1} 	& \textbf{16.3} \\
\cmidrule{2-12}
& \multicolumn{1}{l}{Q2B}    
& 0.5 	& 5.5  	 & 1.7 	  & 0.7   & 0.7 	& 7.8 	& \textbf{12.9} 	& 2.7 	& 6.1 	& 4.3  \\
& \multicolumn{1}{c}{+\textsc{Temp}}  
& \textbf{12.9} 	& \textbf{12.7} 	& \textbf{12.3} 	& \textbf{10.0} 	
& \textbf{10.4} 	& \textbf{13.7}  	& 10.7 	& \textbf{7.8}
& \textbf{9.5} 	    & \textbf{11.1} \\
\cmidrule{2-12}
& \multicolumn{1}{l}{\textsc{BetaE}}    
& 0.9 	& 0.5  	 & 0.4 	  & 0.7   & 0.3 	& 0.7 	& 0.4 	& 1.0 	& 0.3 	& 0.6 \\
& \multicolumn{1}{c}{+\textsc{Temp}}
& \textbf{10.8} 	& \textbf{15.0} 	& \textbf{10.6} 	& \textbf{10.8}
& \textbf{10.1} 	& \textbf{11.3} 	& \textbf{13.6} 	& \textbf{5.5}
& \textbf{13.6} 	& \textbf{11.3} 
\\
\cmidrule{2-12}
& \multicolumn{1}{l}{\textsc{LogicE}}
& 0.7 	& 2.5  	 & 1.1 	  & 1.0   & 0.9 	& 1.0 	& 3.1 	& 0.3 	& 1.6 	& 1.4 \\
& \multicolumn{1}{c}{+\textsc{Temp}}  
& \textbf{15.7} 	& \textbf{17.1} 	& \textbf{15.1} 	& \textbf{14.6}
& \textbf{13.8} 	& \textbf{13.6} 	& \textbf{16.3} 	& \textbf{7.0} 
& \textbf{14.2} 	& \textbf{14.2} \\
\midrule
\multirow{9}*{\rotatebox{90}{NELL-V3}}
& \multicolumn{1}{l}{GQE}    
& 0.3   & 2.3    & 0.6    & 0.4    & 0.1 	& 3.2 	& 4.9 	& 1.8 	& 3.2 	& 1.9 \\
& \multicolumn{1}{c}{+\textsc{Temp}}  
& \textbf{9.6} 	    & \textbf{5.7}   	& \textbf{6.0} 	    & \textbf{7.2} 	
& \textbf{8.1} 	    & \textbf{4.7}  	& \textbf{6.2} 	    & \textbf{4.0}
& \textbf{4.0} 	    & \textbf{6.2} \\ 
\cmidrule{2-12}
& \multicolumn{1}{l}{Q2B}    
& 0.2 	& 2.2  	 & 0.5 	  & 0.3   & 0.2 	& 2.6 	& 4.8 	& 1.8 	& 2.8 	& 1.7  \\
& \multicolumn{1}{c}{+\textsc{Temp}}  
& \textbf{8.0}   	& \textbf{5.6} 	    & \textbf{6.0} 	    & \textbf{7.6} 	
& \textbf{7.1} 	    & \textbf{4.6}  	& \textbf{5.3}   	& \textbf{4.1}
& \textbf{3.2} 	    & \textbf{5.7} \\
\cmidrule{2-12}
& \multicolumn{1}{l}{\textsc{BetaE}}
& 0.4 	& 0.1  	 & 0.1 	  & 0.5   & 0.1 	& 0.2 	& 0.1   & 0.3 
& 0.2 	& 0.2   \\
& \multicolumn{1}{c}{+\textsc{Temp}}  
& \textbf{8.4} 	& \textbf{4.7} 	& \textbf{5.7} 	& \textbf{5.6} 	& \textbf{5.8}
& \textbf{4.1} 	& \textbf{4.5} 	& \textbf{4.5} 	& \textbf{3.0} 	& \textbf{5.1} 
\\
\cmidrule{2-12}
& \multicolumn{1}{l}{\textsc{LogicE}}
& 0.2 	& 0.5  	 & 0.3 	  & 0.1   & 0.3 	& 0.2 	& 0.3 	& 0.2
& 0.4 	& 0.3   \\
& \multicolumn{1}{c}{+\textsc{Temp}}  
& \textbf{10.7} 	& \textbf{5.0} 	& \textbf{5.8} 	& \textbf{7.6} 	& \textbf{8.1} 
& \textbf{5.3} 	    & \textbf{5.5} 	& \textbf{4.7} 	& \textbf{3.4} 	& \textbf{6.2} 
\\
\bottomrule
\end{tabular*}
\caption{Hits@10 results on queries generated from the FB15k-237-V2 and NELL-V3 inductive datasets from GraIL.}
\label{table_2}
\end{table}
\begin{table}[!htp]
\renewcommand\arraystretch{0.50}
\setlength{\tabcolsep}{0.22em}
\centering
\small
\begin{tabular*}{\linewidth}{@{}ccccccccccc@{}}
\toprule
\multicolumn{1}{c}{\textbf{Datasets}} & \textbf{Model} & \textbf{Our model} & \textbf{2in} & \textbf{3in} & \textbf{inp} & \textbf{pin} & \textbf{pni} & \textbf{Avg} \\
\midrule
\multirow{4}{*}{FB15k}      
& \multirow{2}{*}{\textsc{BetaE}}    & None   & 14.3  & 14.7 	& \textbf{11.5} 	& 6.5 	& 12.4
& 11.8 \\
&   & +\textsc{Temp}  & \textbf{15.2} 	& \textbf{15.6} 	& \textbf{11.5} 	
& \textbf{6.8} & \textbf{13.4} 	& \textbf{12.5} \\
\cmidrule{2-9}
& \multirow{2}{*}{\textsc{LogicE}}    & None   & 15.1  & 14.2 	& 12.5 	& 7.1 	& 13.4
& 12.5 \\
&   & +\textsc{Temp}  & \textbf{15.2} 	& \textbf{14.7} 	& \textbf{12.7} 	
& \textbf{7.6} & \textbf{13.7} 	& \textbf{12.8} \\
\midrule
\multirow{4}{*}{FB15k-237}      
& \multirow{2}{*}{\textsc{BetaE}}    & None   & \textbf{5.1}  & 7.9 	& 7.4 	& \textbf{3.6} 	& \textbf{3.4}    & \textbf{5.4} \\
&   & +\textsc{Temp}  & 4.3 	& \textbf{8.0} 	& \textbf{7.6} 	& 3.5   & 2.9 	& 5.3 \\
\cmidrule{2-9}
& \multirow{2}{*}{\textsc{LogicE}}    & None   & 4.9  & 8.2 	& 7.7 	& 3.6 	& 3.5   & 5.6 \\
&   & +\textsc{Temp}  & \textbf{5.4} 	& \textbf{8.7} 	& \textbf{7.9} 	& \textbf{4.0}
& \textbf{3.8} 	& \textbf{6.0} \\
\midrule
\multirow{4}{*}{NELL}      
& \multirow{2}{*}{\textsc{BetaE}}    & None   & \textbf{5.1}  & \textbf{7.8} 	& 10.0 	
& \textbf{3.1} 	& \textbf{3.5}    & \textbf{5.9} \\
&   & +\textsc{Temp}  & \textbf{5.1} 	& 7.5 	& \textbf{10.5} 	& \textbf{3.1}   & 3.3 	
& \textbf{5.9} \\
\cmidrule{2-9}
& \multirow{2}{*}{\textsc{LogicE}}    & None   & 5.3  & 7.5 	& 11.1 	& 3.3 	& 3.8   & 6.2 \\
&   & +\textsc{Temp}  & \textbf{5.4} 	& \textbf{7.6} 	& \textbf{11.3} 	& \textbf{3.4}
& \textbf{3.9} 	& \textbf{6.3} \\
\bottomrule
\end{tabular*}
\caption{MRR Results on the \textsc{BetaE} Datasets for \textsc{BetaE} and \textsc{LogicE} on queries with negation. See appendix  for  results for other query structures.}
\label{table_3}
\end{table}

\paragraph{Inductive Reasoning.} \vic{Table \ref{table_2} shows that the addition of  \textsc{Temp} significantly outperforms all baselines  in the (entity-level) inductive setting, by 12.2\%, 6.8\%, 10.7\%, and 12.8\%, in terms of Hits@10 in the FB15k-237-V2 dataset. 
The main reason is that \vicf{in \textsc{Temp}} entity and relation representations  are mainly characterized by the type information. So,  newly emerging entities or relations can be  captured through their type information, making unnecessary  the coupling between entities and entity set, and  relations and relation set.}

\paragraph{Queries with Negation.} \vic{Table \ref{table_3} shows the results of \textsc{BetaE} and \textsc{LogicE} 
on queries with negation in the \textsc{BetaE} dataset.
\vicf{
The main reason for the small gains is that, unlike \textsc{BetaE} and \textsc{LogicE}, 
\textsc{Temp} does not have specific mechanisms to deal with negation. Specifically, \textsc{Temp} lacks mechanisms to associate  type information to the negation of a relation, i.e., a way to `negate' a type. Boosting queries with negation using type information is left as an interesting future work.}} 

\subsection{Ablation Studies}
\label{ablation_studies}

\begin{figure}[htbp]
    \begin{minipage}[t]{0.45\linewidth}
        \includegraphics[width=0.9\linewidth]{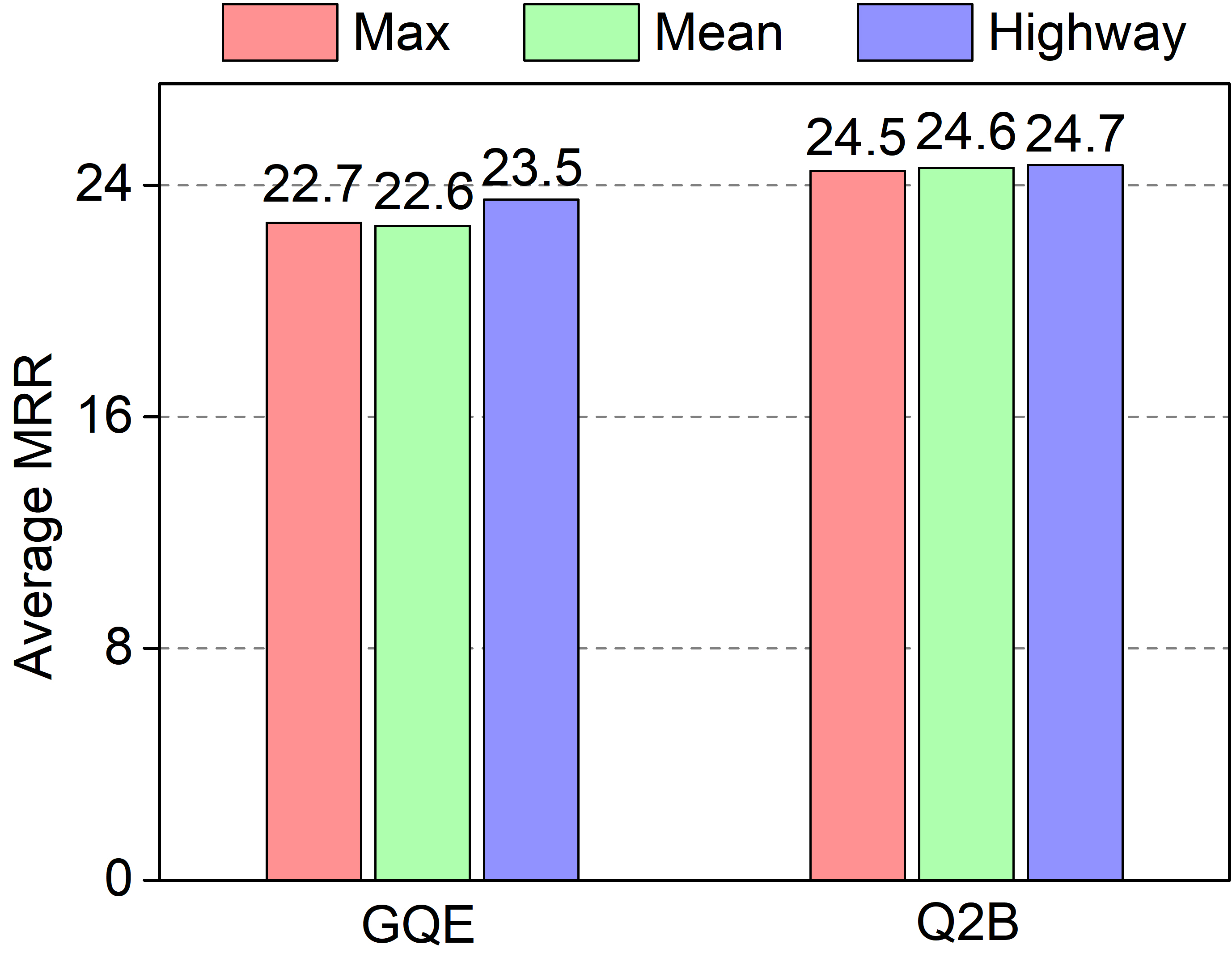} 
        \caption{MRR results with different entity type aggregators.}
        \label{Figure_4}
    \end{minipage}\hfill
    \begin{minipage}[t]{0.45\linewidth}
        \includegraphics[width=0.9\linewidth]{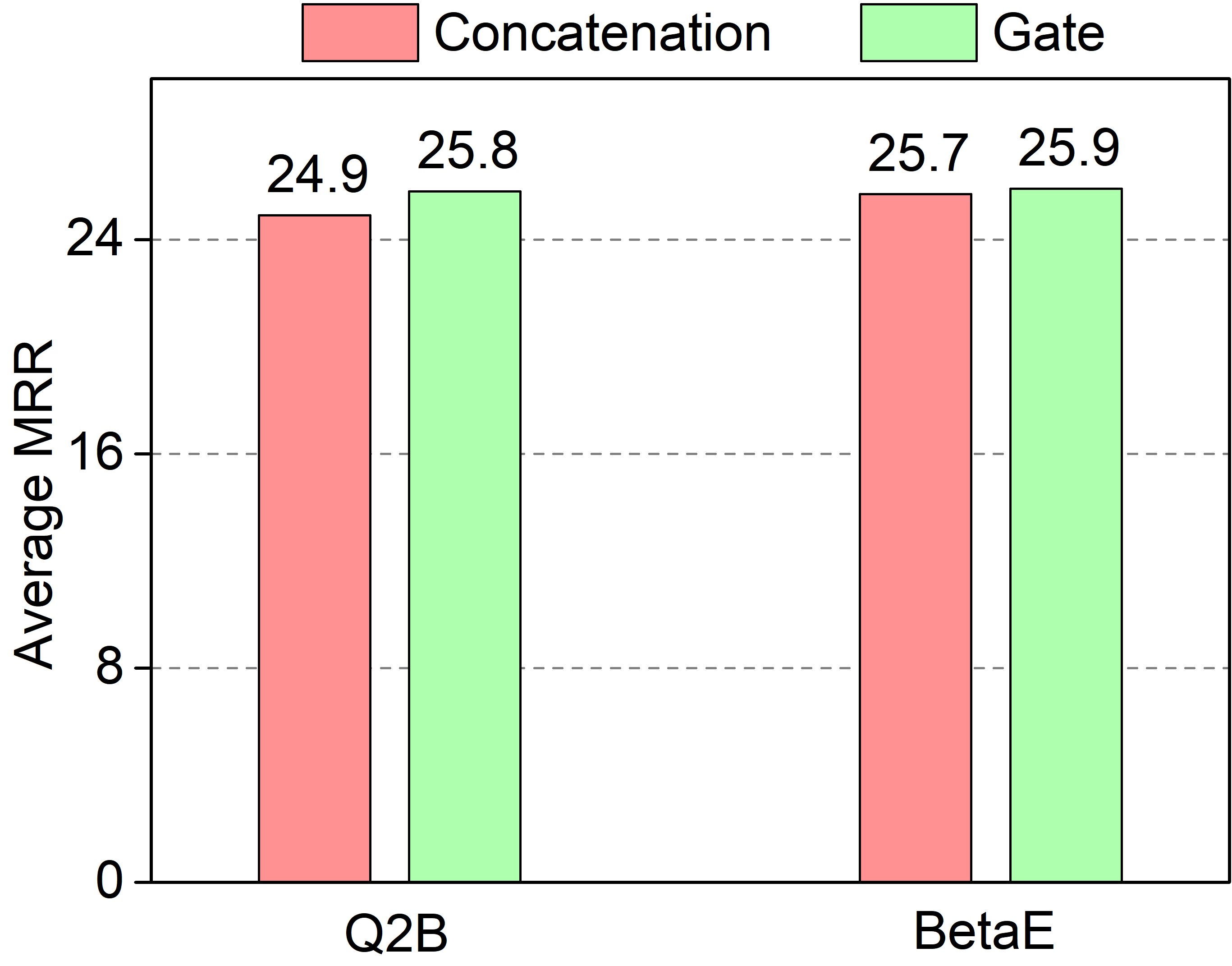} 
        \caption{MRR results with different relation type aggregators.}
        \label{figure_5}
    \end{minipage}
\end{figure}

\vicf{We select   GQE and Q2B, as they  benefit the most by adding type information, 
and conduct ablation experiments on the \zhiwei{three} datasets to study the effect of separately using type-enhancement on entities or relations, see Table \ref{table_4}.  We further study  different implementations of entity and relation  type aggregation models, \zhiwei{see Figure \ref{Figure_4} and Figure \ref{figure_5}.}} 
\begin{table}[!htp]
\renewcommand\arraystretch{0.50}
\setlength{\tabcolsep}{0.65em}
\centering
\scalebox{0.85}{
\begin{tabular*}{\linewidth}{@{}cccccc@{}}
\toprule
\textbf{Models}  & \textbf{TER}    & \textbf{TRR} 
& \textbf{FB15k-237} &\textbf{FB15k} & \textbf{NELL}\\
\midrule
\multirow{4}*{GQE}     
& \tiny \XSolidBold  & \tiny \XSolidBold 
& 19.4      & 32.8      & 19.3 \\
& \tiny \CheckmarkBold  & \tiny \XSolidBold 
& 23.5      & 42.1      & 26.8 \\
& \tiny \XSolidBold  & \tiny \CheckmarkBold 
& 26.3      & 50.8      & 33.2 \\
& \tiny \CheckmarkBold  & \tiny \CheckmarkBold 
& \textbf{28.2}  & \textbf{50.9}      & \textbf{34.5} \\
\midrule
\multirow{4}*{Q2B}     
& \tiny \XSolidBold  & \tiny \XSolidBold 
& 24.2      & 43.4      & 25.7 \\
& \tiny \CheckmarkBold  & \tiny \XSolidBold 
& 24.7      & 44.3      & 29.1 \\
& \tiny \XSolidBold  & \tiny \CheckmarkBold 
& 25.8      & 47.7      & 33.4 \\
& \tiny \CheckmarkBold  & \tiny \CheckmarkBold 
& \textbf{26.8}   & \textbf{49.7}      & \textbf{33.9} \\
\bottomrule
\end{tabular*}
}
\caption{Average MRR results on the Q2B datasets with TER or TRR. See appendix for detailed results.}
\label{table_4}
\end{table}

 \vicf{Type-enhancement on relations is consistently better than on entities, this is  explained by the fact that  enhancing relation representations 
 is more helpful for queries with long chains of existentially quantified variables as it better deals with cascading errors introduced by  relation projections. 
 We also show that using type-enhancement on both entities and relations usually leads to even better performance.} 

\section{Conclusions}
We proposed \textsc{Temp}, a \emph{type-aware plug-and-play} model for  answering FOL queries on incomplete  KGs. 
We experimentally show that \textsc{Temp}  can significantly improve four state-of-the-art models in terms of generalization, deductive and inductive reasoning abilities across three benchmark datasets consistently. 


\clearpage
\section*{Acknowledgments}
This work has been supported by the National Key Research and Development Program of China (No.2020AAA0106100), by the National Natural Science Foundation of China (No.61936012), by the Chang Jiang Scholars Program (J2019032), by a  Leverhulme Trust Research Project Grant (RPG-2021-140), and by the Centre for Artificial Intelligence, Robotics and Human-Machine Systems (IROHMS) at Cardiff University.


\bibliographystyle{named}
\bibliography{ijcai22.bib}

\begin{thebibliography}{}

\bibitem[\protect\citeauthoryear{Arakelyan \bgroup \em et al.\egroup
  }{2021}]{arakelyan2020complex}
Erik Arakelyan, Daniel Daza, Pasquale Minervini, and Michael Cochez.
\newblock Complex query answering with neural link predictors.
\newblock In {\em ICLR}, 2021.

\bibitem[\protect\citeauthoryear{Bordes \bgroup \em et al.\egroup
  }{2013}]{bordes2013translating}
Antoine Bordes, Nicolas Usunier, Alberto Garc{\'{\i}}a{-}Dur{\'{a}}n, Jason
  Weston, and Oksana Yakhnenko.
\newblock Translating embeddings for modeling multi-relational data.
\newblock In {\em NeurIPS}, 2013.

\bibitem[\protect\citeauthoryear{Chen \bgroup \em et al.\egroup
  }{2021}]{Jiajuninductive}
Jiajun Chen, Huarui He, Feng Wu, and Jie Wang.
\newblock Topology-aware correlations between relations for inductive link
  prediction in knowledge graphs.
\newblock In {\em AAAI}, pages 6271--6278, 2021.

\bibitem[\protect\citeauthoryear{Choudhary \bgroup \em et al.\egroup
  }{2021a}]{choudhary2021probabilistic}
Nurendra Choudhary, Nikhil Rao, Sumeet Katariya, Karthik Subbian, and Chandan
  Reddy.
\newblock Probabilistic entity representation model for reasoning over
  knowledge graphs.
\newblock In {\em NeurIPS}, 2021.

\bibitem[\protect\citeauthoryear{Choudhary \bgroup \em et al.\egroup
  }{2021b}]{choudhary2021self}
Nurendra Choudhary, Nikhil Rao, Sumeet Katariya, Karthik Subbian, and Chandan~K
  Reddy.
\newblock Self-supervised hyperboloid representations from logical queries over
  knowledge graphs.
\newblock In {\em WWW}, 2021.

\bibitem[\protect\citeauthoryear{Daza \bgroup \em et al.\egroup
  }{2021}]{daza2021inductive}
Daniel Daza, Michael Cochez, and Paul Groth.
\newblock Inductive entity representations from text via link prediction.
\newblock In {\em WWW}, 2021.

\bibitem[\protect\citeauthoryear{Hamilton \bgroup \em et al.\egroup
  }{2018}]{hamilton2018embedding}
William~L Hamilton, Payal Bajaj, Marinka Zitnik, Dan Jurafsky, and Jure
  Leskovec.
\newblock Embedding logical queries on knowledge graphs.
\newblock In {\em NeurIPS}, 2018.

\bibitem[\protect\citeauthoryear{Luus \bgroup \em et al.\egroup
  }{2021}]{luus2021logic}
Francois Luus, Prithviraj Sen, Pavan Kapanipathi, Ryan Riegel, Ndivhuwo
  Makondo, Thabang Lebese, and Alexander Gray.
\newblock Logic embeddings for complex query answering.
\newblock In {\em NeurIPS}, 2021.

\bibitem[\protect\citeauthoryear{Niu \bgroup \em et al.\egroup
  }{2020}]{niu2020autoeter}
Guanglin Niu, Bo~Li, Yongfei Zhang, Shiliang Pu, and Jingyang Li.
\newblock Autoeter: Automated entity type representation for knowledge graph
  embedding.
\newblock In {\em EMNLP}, 2020.

\bibitem[\protect\citeauthoryear{Pan \bgroup \em et al.\egroup
  }{2016}]{Pan2016}
J.Z. Pan, G.~Vetere, J.M. Gomez-Perez, and H.~Wu.
\newblock {\em {Exploiting Linked Data and Knowledge Graphs for Large
  Organisations}}.
\newblock Springer, 2016.

\bibitem[\protect\citeauthoryear{Pan \bgroup \em et al.\egroup
  }{2021}]{pan2021context}
Weiran Pan, Wei Wei, and Xianling Mao.
\newblock Context-aware entity typing in knowledge graphs.
\newblock In {\em EMNLP}, 2021.

\bibitem[\protect\citeauthoryear{Pan}{2009}]{DBLP:series/ihis/Pan09}
Jeff~Z. Pan.
\newblock Resource description framework.
\newblock In {\em Handbook on Ontologies}, pages 71--90. 2009.

\bibitem[\protect\citeauthoryear{Ren and Leskovec}{2020}]{ren2020beta}
Hongyu Ren and Jure Leskovec.
\newblock Beta embeddings for multi-hop logical reasoning in knowledge graphs.
\newblock In {\em NeurIPS}, 2020.

\bibitem[\protect\citeauthoryear{Ren \bgroup \em et al.\egroup
  }{2020}]{query2box:ren2020}
Hongyu Ren, Weihua Hu, and Jure Leskovec.
\newblock Query2box: Reasoning over knowledge graphs in vector space using box
  embeddings.
\newblock In {\em ICLR}, 2020.

\bibitem[\protect\citeauthoryear{Srivastava \bgroup \em et al.\egroup
  }{2015}]{highway:2015}
Rupesh~Kumar Srivastava, Klaus Greff, and J{\"{u}}rgen Schmidhuber.
\newblock Highway networks.
\newblock {\em arXiv preprint arXiv:1505.00387}, 2015.

\bibitem[\protect\citeauthoryear{Sun \bgroup \em et al.\egroup
  }{2020}]{sun2020faithful}
Haitian Sun, Andrew~O. Arnold, Tania Bedrax{-}Weiss, Fernando Pereira, and
  William~W. Cohen.
\newblock Faithful embeddings for knowledge base queries.
\newblock In {\em NeurIPS}, 2020.

\bibitem[\protect\citeauthoryear{Tai \bgroup \em et al.\egroup
  }{2015}]{tai2015improved}
Kai~Sheng Tai, Richard Socher, and Christopher~D. Manning.
\newblock Improved semantic representations from tree-structured long
  short-term memory networks.
\newblock In {\em ACL}, pages 1556--1566, 2015.

\bibitem[\protect\citeauthoryear{Teru \bgroup \em et al.\egroup
  }{2020}]{teru2020inductive}
Komal Teru, Etienne Denis, and Will Hamilton.
\newblock Inductive relation prediction by subgraph reasoning.
\newblock In {\em ICML}, pages 9448--9457, 2020.

\bibitem[\protect\citeauthoryear{Toutanova and
  Chen}{2015}]{toutanova2015observed}
Kristina Toutanova and Danqi Chen.
\newblock Observed versus latent features for knowledge base and text
  inference.
\newblock In {\em ACL}, pages 57--66, 2015.

\bibitem[\protect\citeauthoryear{Wang \bgroup \em et al.\egroup
  }{2021}]{wang2021relational}
Hongwei Wang, Hongyu Ren, and Jure Leskovec.
\newblock Relational message passing for knowledge graph completion.
\newblock In {\em SIGKDD}, pages 1697--1707, 2021.

\bibitem[\protect\citeauthoryear{Wiharja \bgroup \em et al.\egroup
  }{2020}]{WPKD2020}
Kemas Wiharja, Jeff~Z. Pan, Martin~J. Kollingbaum, and Yu~Deng.
\newblock {Schema Aware Iterative Knowledge Graph Completion}.
\newblock {\em Journal of Web Semantics}, 2020.

\bibitem[\protect\citeauthoryear{Xiong \bgroup \em et al.\egroup
  }{2017}]{xiong2017deeppath}
Wenhan Xiong, Thien Hoang, and William~Yang Wang.
\newblock Deeppath: {A} reinforcement learning method for knowledge graph
  reasoning.
\newblock In {\em EMNLP}, 2017.

\bibitem[\protect\citeauthoryear{Yao \bgroup \em et al.\egroup
  }{2019}]{yao2019kg}
Liang Yao, Chengsheng Mao, and Yuan Luo.
\newblock Kg-bert: Bert for knowledge graph completion.
\newblock {\em arXiv preprint arXiv:1909.03193}, 2019.

\bibitem[\protect\citeauthoryear{Zhang \bgroup \em et al.\egroup
  }{2020}]{zhang2020dcmn}
Shuailiang Zhang, Hai Zhao, Yuwei Wu, Zhuosheng Zhang, Xi~Zhou, and Xiang Zhou.
\newblock Dcmn+: Dual co-matching network for multi-choice reading
  comprehension.
\newblock In {\em AAAI}, 2020.

\bibitem[\protect\citeauthoryear{Zhang \bgroup \em et al.\egroup
  }{2021}]{zhang2021cone}
Zhanqiu Zhang, Jie Wang, Jiajun Chen, Shuiwang Ji, and Feng Wu.
\newblock Cone: Cone embeddings for multi-hop reasoning over knowledge graphs.
\newblock In {\em NeurIPS}, 2021.

\bibitem[\protect\citeauthoryear{Zhao \bgroup \em et al.\egroup
  }{2020}]{zhao2020connecting}
Yu~Zhao, Anxiang Zhang, Ruobing Xie, Kang Liu, and Xiaojie Wang.
\newblock Connecting embeddings for knowledge graph entity typing.
\newblock In {\em ACL}, 2020.

\end{thebibliography}

\clearpage
 \section*{Appendix}
\begin{table*}[htp]
\renewcommand\arraystretch{0.60}
\setlength{\belowcaptionskip}{-0.5cm} 
\setlength{\tabcolsep}{0.8em}
\centering
\begin{tabular*}{\linewidth}{@{}cccccccc@{}}
\toprule
& \textbf{Queries} & \multicolumn{2}{c}{\textbf{Training}} & \multicolumn{2}{c}{\textbf{Validation}} & \multicolumn{2}{c}{\textbf{Test}} \\
\midrule
\textbf{Reasoning Type} & \textbf{Datasets} & \textbf{1p/2p/3p/2i/3i} & \textbf{2in/3in/inp/pin/pni} & \textbf{1p} & \textbf{others} & \textbf{1p} & \textbf{others} \\
\midrule
\multirow{3}*{Deductive Reasoning}
&FB15k    &273,710     &27,371       &59,097      &8,000       &67,016  &8,000 \\
&FB15k-237    &149,689     &14,9,68       &20,101      &5,000  &22,812  &5,000 \\
&NELL995    &107,982     &10,798       &16,927      &4,000       &17,034 &4,000 \\
\midrule
\multirow{2}*{Inductive Reasoning}
&FB15k-237-V2    &9,964  &-  &1,738  &2,000  &791    &1,000 \\
&NELL995-V3      &12,010 &-  &2,197  &2,000  &1,167  &1,500 \\
\bottomrule
\end{tabular*}
\caption{Number of training, validation, and test queries generated for different query structures on \textsc{BetaE} datasets. For Q2B datasets, except 2in/3in/inp/pin/pni, the remaining number of queries is consistent with the \textsc{BetaE} datasets. - denotes that we did not generate the query structures with negation operation for inductive reasoning.}
\label{table_5}
\end{table*}

\subsection*{A Details about Experiments}
\label{detailed_experiments}
\vspace{-0.02cm}
\subsubsection*{A.1  Datasets}
\label{datasets}
For a fair comparison, we use the same datasets and query structures as in Q2B~\cite{query2box:ren2020} and \textsc{BetaE}~\cite{ren2020beta}. 
For testing the inductive reasoning ability, we select GraIL~\cite{teru2020inductive} providing datastes FB15k-237-V2 and NELL995-V3. We  generate  queries over these datasets following~\cite{ren2020beta}. 
The statistics on the number of different queries in different datasets with different reasoning abilities can be found in Table \ref{table_5}.
\vspace{-0.02cm}
\zhiwei{\subsubsection*{A.2  Training Protocol}}
\zhiwei{
We run all the experiments on a single Tesla V100 32G GPU card. All the models are implemented in Pytorch. 
We adopt the original training objective and experimental parameters of each model.
The best hyperparameters are shown in Table \ref{table_6}, note that we use the same parameters before and after adding \textsc{Temp}.}
\vspace{-0.35cm}
\begin{table}[htp]
\renewcommand\arraystretch{0.60}
\setlength{\abovecaptionskip}{-0.05cm}
\setlength{\tabcolsep}{0.40em}
\setlength{\belowcaptionskip}{-0.4cm}
\centering
\begin{tabular*}{\linewidth}{@{}cccccc@{}}
\toprule
model & emb\_dim & lr & batch\_size & neg\_size & margin\\
\toprule
GQE & 800 & 0.0001 & 512 & 128 & 24 \\
Q2B & 400 & 0.0001 & 512 & 128 & 24 \\
\textsc{BETAE} & 400 & 0.0001 & 512 & 128 & 60 \\
\textsc{LOGICE} & 400 & 0.0001 & 512 & 128 & 0.375 \\
\midrule
\end{tabular*}
\caption{ \textcolor{black}{emb\_dim represents the embedding dimension, lr means the learning rate, neg\_size is the negative sample size, and margin means the margin size.
}}
\label{table_6}
\end{table}

\subsubsection*{A.3  Query Structures}
\label{query_structures}
\begin{figure}[ht]
\setlength{\belowcaptionskip}{-0.5cm} 
    \centering
    \includegraphics[width=0.45\textwidth]{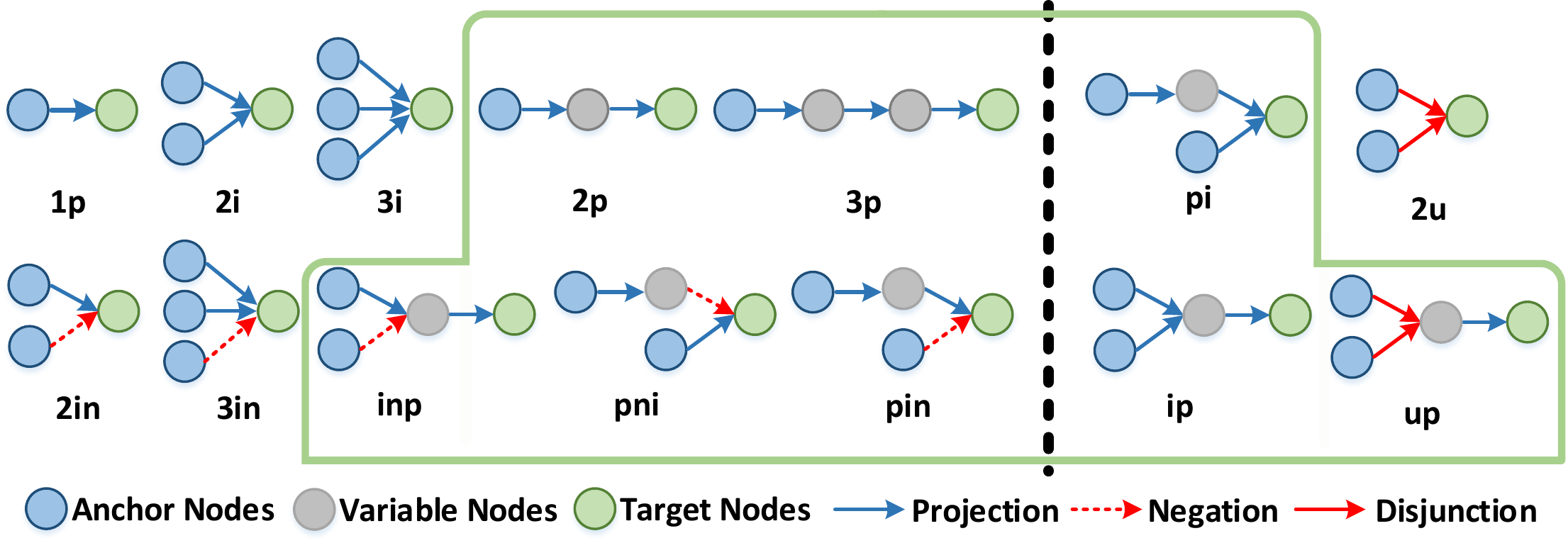}
    \caption{\textbf{Fourteen queries considered in the experiments.} Where `p', `i', `u', and `n' stand for `projection', `intersection', `union', and `negation', respectively.
    }
    \label{fig_2}
\end{figure}

Figure \ref{fig_2} shows  the query structures on the \textsc{BetaE} datasets. The queries on the left
of the black dotted line are used in the training phase. All fourteen queries in Figure \ref{fig_2} are used in both the validation and test phases. The queries within the green solid area are queries containing  existential variable nodes. Unlike the \textsc{BetaE} datasets, the datasets provided by Q2B do not include query structures with negation operations (\emph{2in/3in/inp/pni/pin}).
We refer readers to the original Q2B and \textsc{BetaE} papers for technical details.

\subsubsection*{A.4  Evaluation Metrics}
\label{evaluation_metrics}
The metrics Hit@K and MRR can be defined as follows:
\begin{equation}
\setlength{\abovedisplayskip}{-1pt}
\setlength{\belowdisplayskip}{-3pt}
\textit{Hits}@K(q) = \frac{1}{|A_{q}|}\sum_{v\in A_{q}}{\Phi(\textit{Rank}(v)\leq K)}
\end{equation}
\begin{equation}
\setlength{\belowdisplayskip}{-3pt}
\textit{MRR} = \frac{1}{|A_{q}|}\sum_{v\in A_{q}}{\frac{1}{\textit{Rank}(v)}}
\end{equation}
 $A_{q}$ represents the answer set of \emph{q}, Rank(\emph{v}) denotes as the rank of \emph{v} entity. $\Phi(\cdot)$ is an indicator function, when $x \leq K$, it equals to 1; 0 otherwise. Clearly, higher Hits@K and MRR indicate better prediction performance.

\begin{table*}[htp]
\renewcommand\arraystretch{0.60}
\setlength{\tabcolsep}{1.00em}
\setlength{\belowcaptionskip}{-0.4cm}
\centering
\begin{tabular*}{\linewidth}{@{}cccccccccccc@{}}
\toprule
\multicolumn{1}{c|}{\textbf{Generalization}} & \multicolumn{1}{c|}{\textbf{Method}} & \textbf{1p} & \textbf{2p} & \textbf{3p} & \textbf{2i} & \multicolumn{1}{c|}{\textbf{3i}} & \textbf{pi} & \textbf{ip} & \textbf{2u} & \multicolumn{1}{c|}{\textbf{up}} & \textbf{Avg} \\
\midrule
\multirow{12}*{FB15k}
&GQE   &54.6 	&15.3 	&10.8 	&39.7 	&51.4 	&27.6 	&19.1 	&22.1 	&11.6  &28.0 \\
&\multicolumn{1}{r}{+\textsc{Temp}}   &\textbf{74.9} 	&\textbf{31.4} 	&\textbf{26.0} 	&\textbf{59.3} 	&\textbf{69.4} 	&\textbf{47.3}
&\textbf{35.9} 	&\textbf{47.8} 	&\textbf{27.4} 	&\textbf{46.6} \\
\cmidrule{2-12}
&Q2B   &68.0 	&21.0 	&14.2 	&55.1 	&66.5 	&39.4 	&26.1 	
&35.1  &16.7 	&38.0 \\
&\multicolumn{1}{r}{+\textsc{Temp}}   &\textbf{74.8} 	&\textbf{25.6} 	&\textbf{22.3} 	&\textbf{61.7} 	&\textbf{72.6} 	&\textbf{43.7}
&\textbf{29.0} 	&\textbf{44.1} 	&\textbf{22.5} 	&\textbf{44.0} \\
\cmidrule{2-12}
&\textsc{BetaE}   &65.1 	&25.7 	&24.7 	&55.8 	&66.5 	&43.9 	&28.1 	
&40.1 	&25.2 	&41.6 \\
&\multicolumn{1}{r}{+\textsc{Temp}}   &\textbf{70.3} 	&\textbf{28.9}
&\textbf{25.8} 	&\textbf{58.2} 	&\textbf{68.4}  &\textbf{45.8}
&\textbf{32.2} 	&\textbf{44.3} 	&\textbf{27.3} 	&\textbf{44.6} \\
\cmidrule{2-12}
&\textsc{LogicE}   &72.3 	&29.8 	&26.2 	&56.1 	&66.3 	&42.7 	&32.6 	
&43.4 	&27.5 	&44.1 \\
&\multicolumn{1}{r}{+\textsc{Temp}}   &\textbf{74.9} 	&\textbf{30.9} 	&\textbf{27.3}  &\textbf{58.3} 	&\textbf{68.2} 	&\textbf{43.8}
&\textbf{33.8} 	&\textbf{46.1} 	&\textbf{28.9} 	&\textbf{45.8} \\

\midrule
\multirow{12}*{FB15k-237}
&GQE   &35.0 	&7.2 	&5.3 	&23.3 	&34.6 	&16.5 	&10.7 	
&8.2   &5.7 	&16.3 \\
&\multicolumn{1}{r}{+\textsc{Temp}}   &\textbf{42.9} 	&\textbf{12.3}
&\textbf{10.1} 	&\textbf{34.4} 	&\textbf{47.6} 	&\textbf{26.0}
&\textbf{15.1} 	&\textbf{15.1} 	&\textbf{10.1} 	&\textbf{23.7} \\
\cmidrule{2-12}
&Q2B   &40.6 	&9.4 	&6.8 	&29.5 	&42.3 	&21.2 	&\textbf{12.6} 	
&11.3 	&7.6 	&20.1 \\
&\multicolumn{1}{r}{+\textsc{Temp}}   &\textbf{40.9} 	&\textbf{11.0}
&\textbf{9.2} 	&\textbf{33.7} 	&\textbf{48.2} 	&\textbf{21.4}
&12.3 	&\textbf{12.9} 	&\textbf{9.1} 	&\textbf{22.1}  \\
\cmidrule{2-12}
&\textsc{BetaE}   &39.0 	&10.9 	&10.0 	&28.8 	&42.5 	&22.4 	&12.6 	
&12.4    &9.7 	&20.9 \\
&\multicolumn{1}{r}{+\textsc{Temp}}   &\textbf{39.9} 	&\textbf{11.8}
&\textbf{10.5} 	&\textbf{32.6} 	&\textbf{46.7} 	&\textbf{24.9}
&\textbf{13.6} 	&\textbf{12.5} 	&\textbf{10.2} 	&\textbf{22.5} \\
\cmidrule{2-12}
&\textsc{LogicE}   &41.3 	&11.8 	&10.4 	&31.4 	&43.9 	&23.8 	&14.0 	
&13.4 	&10.2 	&22.3 \\
&\multicolumn{1}{r}{+\textsc{Temp}}   &\textbf{41.9} 	&\textbf{13.0}
&\textbf{11.2} 	&\textbf{32.4} 	&\textbf{45.4} 	&\textbf{25.5} 	
&\textbf{15.2} 	&\textbf{14.6} 	&\textbf{10.9} 	&\textbf{23.3} \\

\midrule
\multirow{12}*{NELL}
&GQE   &32.8 	&11.9 	&9.6 	&27.5 	&35.2 	&18.4 	&14.4 	
&8.5   &8.8 	&18.6 \\
&\multicolumn{1}{r}{+\textsc{Temp}}   &\textbf{57.7} 	&\textbf{17.2}
&\textbf{14.1} 	&\textbf{40.6} 	&\textbf{49.9} 	&\textbf{27.0} 
&\textbf{18.5} 	&\textbf{15.9} 	&\textbf{11.6} 	&\textbf{28.0} \\
\cmidrule{2-12}
&Q2B   &42.2 	&14.0 	&11.2 	&33.3 	&44.5 	&\textbf{22.4} 	&\textbf{16.8}  &11.3  &\textbf{10.3} 	&22.9 \\
&\multicolumn{1}{r}{+\textsc{Temp}}   &\textbf{56.5} 	&\textbf{15.0}
&\textbf{12.9} 	&\textbf{40.8} 	&\textbf{52.0} 	&21.1	
&16.0 	&\textbf{14.2} 	&9.4 	&\textbf{26.4}  \\
\cmidrule{2-12}
&\textsc{BetaE}   &53.0 	&13.0 	&11.4 	&37.6 	&47.5 	&\textbf{24.1} 	&14.3 	
&12.2    &8.5 	&24.6 \\
&\multicolumn{1}{r}{+\textsc{Temp}}   &\textbf{54.1} 	&\textbf{14.2}
&\textbf{12.4} 	&\textbf{38.1} 	&\textbf{48.9} 	&23.9
&\textbf{16.0} 	&\textbf{12.8} 	&\textbf{9.2} 	&\textbf{25.5} \\
\cmidrule{2-12}
&\textsc{LogicE}   &58.3 	&17.7 	&15.4 	&40.5 	&50.4 	&27.3 	&19.2 	
&15.9     &12.7 	&28.6 \\
&\multicolumn{1}{r}{+\textsc{Temp}}   &\textbf{58.4} 	&\textbf{18.1}
&\textbf{14.9} 	&\textbf{40.7} 	&\textbf{50.6} 	&\textbf{27.9}
&\textbf{19.1} 	&\textbf{16.7} 	&\textbf{12.8} 	&\textbf{28.8} \\

\bottomrule
\end{tabular*}
\caption{Detailed MRR results on the \textsc{BetaE}'s datasets testing generalization.}
\label{table_7}
\end{table*}

\begin{table*}[htp]
\renewcommand\arraystretch{0.60}
\setlength{\tabcolsep}{0.90em}
\setlength{\belowcaptionskip}{-0.4cm}
\small
\centering
\begin{tabular*}{\linewidth}{@{}cccccccccccccc@{}}
\toprule
\multicolumn{1}{c|}{\textbf{Datasets}} & \multicolumn{1}{c|}{\textbf{Methods}} & \textbf{TER} & \textbf{TRR} & \textbf{1p} & \textbf{2p} & \textbf{3p} & \textbf{2i} & \multicolumn{1}{c|}{\textbf{3i}} & \textbf{pi} & \textbf{ip} & \textbf{2u} & \multicolumn{1}{c|}{\textbf{up}} & \textbf{Avg} \\
\midrule
\multirow{12}*{FB15k}
&\multirow{5}*{GQE}
&\tiny \XSolidBold  &\tiny \XSolidBold  &55.1 	&30.7 	&22.5 	&38.6 	&49.3 	&24.9 	&14.6 	&34.3 	&25.1 	&32.8 \\
&   &\tiny \CheckmarkBold  & \tiny \XSolidBold   &65.3 	&38.1 	&29.4 	
&52.1 	&64.3 	&35.5 	&17.9 	&44.6 	&31.4 	&42.1 \\
&   &\tiny \XSolidBold  &\tiny \CheckmarkBold  &74.2 	&49.8 	&\textbf{42.0}  &56.2 	&66.6 	&44.3 	&27.3 	&59.1 	&\textbf{37.7} 	&50.8 \\
&   &\tiny \CheckmarkBold  &\tiny \CheckmarkBold  &\textbf{75.2} 	
&\textbf{50.4} 	&\textbf{42.0}  &\textbf{56.9} 	&\textbf{67.1} 	
&\textbf{44.5} 	&\textbf{28.2} 	&\textbf{59.5} 	&34.0 	&\textbf{50.9} \\ \cmidrule{2-14}
&\multirow{5}*{Q2B}
&\tiny \XSolidBold  &\tiny \XSolidBold  &68.3 	&39.0 	&28.5 	&52.2 	&64.0 	&37.4 	&20.1 	&50.6 	&30.1 	&43.4 \\
&   &\tiny \CheckmarkBold  & \tiny \XSolidBold   &72.4 	&37.3 	&28.6 	&55.7 	&67.7 	&36.0 	&17.4 	&57.1 	&26.3 	&44.3 \\
&   &\tiny \XSolidBold  &\tiny \CheckmarkBold  &69.9 	&\textbf{46.1} 	&\textbf{38.8} 	&56.2 	&68.6 	&41.8 	&\textbf{25.6} 	&52.7 	&\textbf{29.3}  &47.7  \\
&   &\tiny \CheckmarkBold  &\tiny \CheckmarkBold  &\textbf{75.6} 	&45.5	&38.4 	&\textbf{60.0} 	&\textbf{71.0} 	&\textbf{43.1} 	&23.6 	&\textbf{62.7} 	&27.4 	&\textbf{49.7}  \\
\midrule

\multirow{12}*{FB15k-237}
&\multirow{5}*{GQE}
&\tiny \XSolidBold  &\tiny \XSolidBold  &35.0 	&19.2 	&14.4 	&22.1 	&31.4 	&14.5 	&8.8 	&14.8 	&14.5 	&19.4 \\
&   &\tiny \CheckmarkBold  & \tiny \XSolidBold  &40.3 	&22.2 	&17.5 	&27.0 	&38.3 	&17.3 	&9.9 	&21.1 	&18.1 	&23.5 \\
&   &\tiny \XSolidBold  &\tiny \CheckmarkBold   &\textbf{43.9} 	&26.8 	&22.5 	&28.2 	&37.6 	&20.4 	&11.9 	&24.9 	&\textbf{20.9} 	&26.3 \\
&   &\tiny \CheckmarkBold  &\tiny \CheckmarkBold  &43.0 &\textbf{27.7} 	&\textbf{23.4} 	&\textbf{32.6} 	&\textbf{44.4} 	&\textbf{23.4} 	&\textbf{13.2} 	&\textbf{26.5} 	&20.0 	&\textbf{28.2} \\
\cmidrule{2-14}
&\multirow{5}*{Q2B}
&\tiny \XSolidBold  &\tiny \XSolidBold  &40.7 	&23.1 	&18.1 	&27.9 	&38.9 	&19.2 	&10.9 	&21.1 	&\textbf{17.8} 	&24.2 \\
&   &\tiny \CheckmarkBold  & \tiny \XSolidBold   &42.1 	&23.1 	&17.6 	&29.6 	&41.3 	&18.1 	&9.9 	&22.7 	&17.7 	 &24.7 \\
&   &\tiny \XSolidBold  &\tiny \CheckmarkBold    &\textbf{42.2} 	&\textbf{26.0} 	&21.6 	&29.6 	&42.0 	&18.6 	&\textbf{12.2} 	&22.1 	&\textbf{17.8} 	 &25.8  \\
&   &\tiny \CheckmarkBold  &\tiny \CheckmarkBold &41.2 	&25.8 	&\textbf{21.9} 	&\textbf{33.1} 	&\textbf{45.0} 	&\textbf{21.0} 	&11.4 	&\textbf{24.4} 	&17.7  	 &\textbf{26.8}  \\
\midrule

\multirow{12}*{NELL}
&\multirow{5}*{GQE}
&\tiny \XSolidBold  &\tiny \XSolidBold  &31.3 	&18.2 	&16.8 	&22.4 	&30.4 	&15.8 	&10.0 	&16.8 	&12.1 	&19.3 \\
&   &\tiny \CheckmarkBold  & \tiny \XSolidBold  &51.2 	&21.9 	&21.7 	&28.5 	&40.6 	&17.5 	&9.2 	&35.1 	&15.6 	&26.8 \\
&   &\tiny \XSolidBold  &\tiny \CheckmarkBold   &55.5 	&31.8 	&31.7 	&33.6 	&46.1 	&23.0 	&14.1 	&38.9 	&24.4 	&33.2 \\
&   &\tiny \CheckmarkBold  &\tiny \CheckmarkBold  &\textbf{56.7} &\textbf{33.6} 	&\textbf{32.4} 	&\textbf{35.5} 	&\textbf{47.2} 	&\textbf{23.6} 	&\textbf{14.8} 	&\textbf{41.6} 	&\textbf{24.8} 	&\textbf{34.5} \\
\cmidrule{2-14}
&\multirow{5}*{Q2B}
&\tiny \XSolidBold  &\tiny \XSolidBold  &41.8 	&23.1 	&21.2 	&28.9 	&41.7 	&\textbf{19.7} 	&12.2 	&26.9 	&15.5 	&25.7 \\
&   &\tiny \CheckmarkBold  & \tiny \XSolidBold   &54.4 	&22.9 	&23.0 	&33.7 	&47.4 	&15.9 	&10.1 	&38.7 	&15.8 	&29.1 \\
&   &\tiny \XSolidBold  &\tiny \CheckmarkBold    &55.9 	&31.1 	&\textbf{31.3} 	&34.8 	&48.9 	&19.4 	&\textbf{14.0} 	&41.2 	&23.6 	&33.4  \\
&   &\tiny \CheckmarkBold  &\tiny \CheckmarkBold &\textbf{57.1} 	&\textbf{31.9} 	&\textbf{31.3} 	&\textbf{36.6} 	&\textbf{49.5} 	&19.4 	&13.5 	&\textbf{41.7} 	&\textbf{23.8} 	&\textbf{33.9}  \\

\bottomrule
\end{tabular*}
\caption{Detailed MRR results on the Q2B datasets with TER or TRR.}
\label{table_8}
\end{table*}

\subsection*{B  Design Alternatives}
\label{section_design_alternatives}
In our ablation experiments, we test \textsc{Temp} with the following alternative models. In particular, when aggregating the type of an entity, we propose two alternatives for aggregator, instead of the highway aggregator in Eqs. (\ref{highway_agg_1}), (\ref{highway_agg_2}), and (\ref{highway_agg_3}). \\
\vspace{-0.4cm}
\paragraph{Mean aggregator.}
\begin{equation}
\setlength{\abovedisplayskip}{-3pt}
\setlength{\belowdisplayskip}{-3pt}
\widetilde{\mathcal{H}_{s}^{K}} = \frac{1}{n}\sum_{i} \mathcal{H}_{s_{i}}^{K}
\end{equation}
Here, the mean operation does not require a loop operation, so the value of \emph{K} is 0. $\mathcal{H}_{s_{i}}^{K}\in\mathbb{R}^{d\times 1}$ represents the \emph{i-th} component of the vector $\mathcal{H}_{s}^{K}\in\mathbb{R}^{d\times n}$. \\
\vspace{-0.4cm}
\paragraph{Max  aggregator.}
\begin{equation}
\setlength{\abovedisplayskip}{-3pt}
\setlength{\belowdisplayskip}{-1pt}
\widetilde{\mathcal{H}_{s}^{K}} = \textit{Max}(\mathcal{H}_{s_{i}}^{K})
\end{equation}
$\textit{Max}(\cdot)$ represents the element-wise maximum operation among the entity's type information. \emph{K} also takes 0.\\
\vspace{-0.4cm}
\paragraph{Concatenation-based intersection.} Inspired by ~\cite{liu2016learning} and ~\cite{reimers2019sentence}, we apply an interactive concatenation to pair of the representations $\mathcal{G}^{er}$ and $\mathcal{G}^{es}$ in Eqs.(\ref{dcmn_1}) and (\ref{dcmn_2}), and then perform a linear layer operation. The interactive concatenation can be specified as:
\begin{equation}
\setlength{\abovedisplayskip}{-2pt}
\setlength{\belowdisplayskip}{-2pt}
\widetilde{\mathcal{G}^{e}} = W[\mathcal{G}^{er}, \mathcal{G}^{es}, \mathcal{G}^{er}+\mathcal{G}^{es}, \mathcal{G}^{er}\times\mathcal{G}^{es}] + b,
\end{equation}
where $+$ and $\times$ represent the element-wise addition and multiplication between two matrices $\mathcal{G}^{er}$ and $\mathcal{G}^{es}$, respectively. $[\cdot, \cdot, \cdot, \cdot]$ is used to concatenate the vector in row level.



\subsection*{C  Additional Results}
\begin{enumerate}
\item Table \ref{table_7} shows  detailed MRR results on the \textsc{BetaE} datasets, testing the generalization ability.

\item Table \ref{table_8} shows  detailed MRR results on the Q2B datasets with TER or TRR used separately  or together.

\item Table \ref{table_9} shows  detailed Hits@3 results on the Q2B datasets about the generalization and deductive reasoning ability.

\end{enumerate}


\begin{table*}[!htp]
\renewcommand\arraystretch{0.60}
\setlength{\tabcolsep}{0.95em}
\centering
\begin{tabular*}{\linewidth}{@{}cccccccccccc@{}}
\toprule
\textbf{Generalization} & \multicolumn{1}{c|}{\textbf{Method}} & \textbf{1p} & \textbf{2p} & \textbf{3p} & \textbf{2i} & \multicolumn{1}{c|}{\textbf{3i}} & \textbf{pi} & \textbf{ip} & \textbf{2u} & \multicolumn{1}{c|}{\textbf{up}} & \textbf{Avg} \\
\midrule
\multirow{12}*{FB15k}
&GQE   &71.7 	&36.1 	&25.9 	&47.5 	&60.8 	&30.0 	&15.8 	&45.2 	&28.1 	&40.1 \\
&\multicolumn{1}{r}{+\textsc{Temp}}   &\textbf{83.0} 	&\textbf{54.6} 	&\textbf{46.0} 	&\textbf{64.1} 	&\textbf{74.8} 	&\textbf{50.0} 	&\textbf{30.8} 	&\textbf{68.7}
&\textbf{37.3} 	&\textbf{56.6} \\
\cmidrule{2-12}
&Q2B   &82.1 	&43.0 	&31.7 	&62.7 	&74.5 	&44.4 	&22.4 	&66.7 	&\textbf{33.5} 	&51.2 \\
&\multicolumn{1}{r}{+\textsc{Temp}}   &\textbf{84.0} 	&\textbf{49.8} 	&\textbf{42.2} 	&\textbf{67.4} 	&\textbf{77.9} 	&\textbf{48.3} 	&\textbf{26.1} 	&\textbf{72.7}
&30.0 	&\textbf{55.4} \\
\cmidrule{2-12}
&\textsc{BetaE}   &75.0 	&45.1 	&40.0 	&62.2 	&\textbf{75.4} 	&47.1 	&22.3 	&58.9 	&29.6 	&50.6 \\
&\multicolumn{1}{r}{+\textsc{Temp}}   &\textbf{79.5} 	&\textbf{50.2} 	&\textbf{44.3} 	&\textbf{63.7} 	&74.4 	&\textbf{48.8} 	&\textbf{27.4} 	&\textbf{64.5}
&\textbf{30.0} 	&\textbf{53.6} \\
\cmidrule{2-12}
&\textsc{LogicE}   &80.5 	&50.5 	&45.8 	&62.2 	&72.5 	&47.5 	&28.3 	&63.9 	&36.8 	&54.2 \\
&\multicolumn{1}{r}{+\textsc{Temp}}   &\textbf{83.2} 	&\textbf{51.5} 	&\textbf{46.4} 	&\textbf{64.0} 	&\textbf{73.8} 	&\textbf{47.8} 	&\textbf{28.6} 	&\textbf{69.0}
&\textbf{36.9} 	&\textbf{55.7} \\
\midrule

\multirow{12}*{FB15k-237}
&GQE   &41.3 	&21.5 	&15.2 	&26.5 	&38.5 	&16.7 	&8.8 	&17.1 	&15.8 	&22.4 \\
&\multicolumn{1}{r}{+\textsc{Temp}}   &\textbf{47.6} 	&\textbf{29.6} 	&\textbf{24.7} 	&\textbf{36.3} 	&\textbf{48.4} 	&\textbf{25.5} 	&\textbf{13.4} 	&\textbf{30.2}
&\textbf{21.0} 	&\textbf{30.7} \\
\cmidrule{2-12}
&Q2B   &\textbf{47.1} 	&24.9 	&19.4 	&33.2 	&46.4 	&21.8 	&11.3 	&25.3 	&\textbf{19.3} 	&27.6 \\
&\multicolumn{1}{r}{+\textsc{Temp}}   &45.7  	&\textbf{27.8} 	&\textbf{23.4} 	&\textbf{36.9} 	&\textbf{49.6} 	&\textbf{22.9} 	&\textbf{11.7} 	&\textbf{27.6}
&18.9 	&\textbf{29.4} \\
\cmidrule{2-12}
&\textsc{BetaE}   &42.6 	&25.4 	&21.6 	&30.2 	&43.3 	&20.7 	&9.2 	&24.2 	&\textbf{18.3} 	&26.2 \\
&\multicolumn{1}{r}{+\textsc{Temp}}   &\textbf{43.3} 	&\textbf{27.2} 	&\textbf{22.7} 	&\textbf{35.3} 	&\textbf{47.5} 	&\textbf{24.3} 	&\textbf{10.6} 	&\textbf{26.7}
&18.2 	&\textbf{28.4} \\
\cmidrule{2-12}
&\textsc{LogicE}   &45.6 	&27.8 	&24.1 	&34.7 	&46.5 	&23.5 	&12.0 	&27.1 	&20.8 	&29.1 \\
&\multicolumn{1}{r}{+\textsc{Temp}}   &\textbf{46.6} 	&\textbf{29.2} 	&\textbf{25.0} 	&\textbf{35.8} 	&\textbf{47.9} 	&\textbf{24.9} 	&\textbf{13.5} 	&\textbf{28.7}
&\textbf{21.0} 	&\textbf{30.3} \\
\midrule

\multirow{12}*{NELL}
&GQE   &42.7 	&21.6 	&19.3 	&27.0 	&37.4 	&17.7 	&10.3 	&21.7 	&13.4 	&23.5 \\
&\multicolumn{1}{r}{+\textsc{Temp}}   &\textbf{62.5} 	&\textbf{36.6} 	&\textbf{35.2} 	&\textbf{40.2} 	&\textbf{52.6} 	&\textbf{25.6} 	&\textbf{15.8} 	&\textbf{48.2}  &\textbf{27.4} 	&\textbf{38.2} \\
\cmidrule{2-12}
&Q2B   &56.2 	&27.1 	&24.1 	&34.7 	&49.2 	&\textbf{21.8} 	&12.7 	&37.5 	&16.5  &31.1 \\
&\multicolumn{1}{r}{+\textsc{Temp}}   &\textbf{62.5}  	&\textbf{34.3} 	&\textbf{34.2} 	&\textbf{41.0} 	&\textbf{55.2} 	&20.9 	&\textbf{14.1} 	&\textbf{47.7}  &\textbf{26.2} 	&\textbf{37.3} \\
\cmidrule{2-12}
&\textsc{BetaE}   &58.4 	&29.1 	&30.7 	&35.2 	&48.4 	&22.5 	&10.5 &\textbf{44.5} 	&\textbf{21.2} 	&33.4 \\
&\multicolumn{1}{r}{+\textsc{Temp}}   &\textbf{58.7} 	&\textbf{31.7} 	&\textbf{32.7} 	&\textbf{37.3} 	&\textbf{51.3} 	&\textbf{22.8} 	
&\textbf{11.7} 	&43.7   &20.6 	&\textbf{34.5} \\
\cmidrule{2-12}
&\textsc{LogicE}   &\textbf{64.3} 	&35.9 	&\textbf{36.0} 	&41.1 	&54.8 	
&\textbf{26.8} 	&14.7
&51.0 	&27.6 	&\textbf{39.1} \\
&\multicolumn{1}{r}{+\textsc{Temp}}   &\textbf{64.3} 	&\textbf{36.2} 	&35.9 	&\textbf{41.2} 	&\textbf{54.9} 	&25.4 	&\textbf{15.5} 	&\textbf{51.1}  &\textbf{27.9} 	&\textbf{39.1} \\

\midrule
\textbf{Deductive Reasoning} & \multicolumn{1}{c|}{\textbf{Method}} & \textbf{1p} & \textbf{2p} & \textbf{3p} & \textbf{2i} & \multicolumn{1}{c|}{\textbf{3i}} & \textbf{pi} & \textbf{ip} & \textbf{2u} & \multicolumn{1}{c|}{\textbf{up}} & \textbf{Avg} \\
\midrule
\multirow{12}*{FB15k}
&GQE   &73.8 	&40.5 	&32.1 	&49.8 	&64.7 	&36.1 	&18.9 	&47.2 	&30.4 	&43.7 \\
&\multicolumn{1}{r}{+\textsc{Temp}}   &\textbf{92.8} 	&\textbf{71.5} 	&\textbf{62.0} 	&\textbf{74.6} 	&\textbf{83.8} 	&\textbf{64.4} 	&\textbf{48.0} 	&\textbf{85.5}  &\textbf{60.0} 	&\textbf{71.4} \\
\cmidrule{2-12}
&Q2B   &68.0 	&39.4 	&32.7 	&48.5 	&65.3 	&32.9 	&16.2 	&61.4
&28.9  &43.7 \\
&\multicolumn{1}{r}{+\textsc{Temp}}   &\textbf{92.8} 	&\textbf{67.1} 	&\textbf{57.3} 	&\textbf{79.2} 	&\textbf{87.6} 	&\textbf{62.8} 	&\textbf{39.8} 	&\textbf{87.3}  &\textbf{52.4} 	&\textbf{69.6} \\
\cmidrule{2-12}
&\textsc{BetaE}   &83.2 	&57.3 	&51.0 	&\textbf{71.1} 	&\textbf{81.4} 	
&\textbf{56.9} 	&32.7 	&70.4 	&41.0 	&\textbf{60.6} \\
&\multicolumn{1}{r}{+\textsc{Temp}}   &\textbf{84.0} 	&\textbf{58.8} 	&\textbf{51.8} 	&68.7 	&78.4 	&55.6 	&\textbf{34.8} 	&\textbf{71.6}
&\textbf{41.9} 	&\textbf{60.6} \\
\cmidrule{2-12}
&\textsc{LogicE}   &88.4 	&64.0 	&57.9 	&70.8 	&80.6 	&59.0 	&41.0 	&76.6 &51.0 	  &65.5 \\
&\multicolumn{1}{r}{+\textsc{Temp}}   &\textbf{91.0} 	&\textbf{64.7} 	&\textbf{58.7} 	&\textbf{72.3} 	&\textbf{81.4} 	&\textbf{60.5} 	&\textbf{42.2} 	&\textbf{81.2}  &\textbf{51.5} 	&\textbf{67.1} \\

\midrule
\multirow{12}*{FB15k-237}
&GQE   &56.4 	&30.1 	&24.5 	&35.9 	&51.2 	&25.1 	&13.0 	&25.8 	&22.0  &31.6 \\
&\multicolumn{1}{r}{+\textsc{Temp}}   &\textbf{76.3} 	&\textbf{48.6} 	&\textbf{39.0} 	&\textbf{49.7} 	&\textbf{60.4} 	&\textbf{36.9} 	&\textbf{22.1} 	&\textbf{59.0}  &\textbf{36.3} 	&\textbf{47.6} \\
\cmidrule{2-12}
&Q2B   &58.5 	&34.3 	&28.1 	&44.7 	&62.1 	&23.9 	&11.7 	
&40.5  &22.0    &36.2 \\
&\multicolumn{1}{r}{+\textsc{Temp}}   &\textbf{87.2} 	&\textbf{59.6} 	&\textbf{47.9} 	&\textbf{67.2} 	&\textbf{72.7} 	&\textbf{49.1} 	&\textbf{29.8} 	&\textbf{78.6}  &\textbf{43.4} 	&\textbf{59.5} \\
\cmidrule{2-12}
&\textsc{BetaE}   &77.9 	&52.6 	&44.5 	&59.0 	&67.8 	&42.2 	&23.5 	&63.7 	&35.1 	 &51.8 \\
&\multicolumn{1}{r}{+\textsc{Temp}}   &\textbf{84.7} 	&\textbf{58.3} 	&\textbf{49.4} 	&\textbf{62.3} 	&\textbf{68.8} 	&\textbf{45.3} 	&\textbf{28.5} 	&\textbf{74.5}  &\textbf{41.4} 	&\textbf{57.0} \\
\cmidrule{2-12}
&\textsc{LogicE}   &81.5 	&54.2 	&46.0 	&58.1 	&67.1 	&44.0 	&28.5
&66.6     &40.8 	&54.1 \\
&\multicolumn{1}{r}{+\textsc{Temp}}   &\textbf{84.5} 	&\textbf{59.8} 	&\textbf{51.9} 	&\textbf{59.3} 	&\textbf{68.1} 	&\textbf{47.0} 	&\textbf{33.4} 	&\textbf{70.8}  &\textbf{45.4} 	&\textbf{57.8} \\

\midrule
\multirow{12}*{NELL}
&GQE   &72.8 	&58.0 	&55.2 	&45.9 	&57.3 	&34.2 	&24.8 	
&59.0  &40.7    &49.8 \\
&\multicolumn{1}{r}{+\textsc{Temp}}   &\textbf{93.3} 	&\textbf{84.1} 	&\textbf{73.3} 	&\textbf{73.4} 	&\textbf{81.4} 	&\textbf{60.8} 	&\textbf{45.8} 	&\textbf{90.3}  &\textbf{76.8} 	&\textbf{75.5} \\
\cmidrule{2-12}
&Q2B   &83.9 	&57.7 	&47.8 	&49.9 	&66.3 	&29.6 	&19.9 	
&73.7  &31.0    &51.1 \\
&\multicolumn{1}{r}{+\textsc{Temp}}   &\textbf{98.3} 	&\textbf{95.4} 	&\textbf{86.0} 	&\textbf{92.2} 	&\textbf{95.2} 	&\textbf{86.1} 	&\textbf{69.5} 	&\textbf{98.5}  &\textbf{92.8} 	&\textbf{90.4} \\
\cmidrule{2-12}
&\textsc{BetaE}   &94.3 	&88.2 	&76.2 	&\textbf{84.0} 	&\textbf{90.2}
&68.8    &46.6 	 &92.5 	&81.4 	&80.2 \\
&\multicolumn{1}{r}{+\textsc{Temp}}   &\textbf{95.1} 	&\textbf{89.8} 	&\textbf{79.5} 	&83.3 	&89.3 	&\textbf{69.0} 	&\textbf{51.1} 	&\textbf{93.0}  &\textbf{83.3} 	&\textbf{81.5} \\
\cmidrule{2-12}
&\textsc{LogicE}   &96.2 	&90.7 	&\textbf{84.1} 	&\textbf{84.1}  &\textbf{89.5} 	&\textbf{76.0} 	&\textbf{65.2}  &94.7    
&\textbf{87.1} 	&\textbf{85.3} \\
&\multicolumn{1}{r}{+\textsc{Temp}}   &\textbf{96.5} 	&\textbf{91.0} 	&83.2 	&82.7 	&88.9 	&72.6 	&64.1 	&\textbf{95.2}  &86.9 	&84.6 \\

\bottomrule
\end{tabular*}
\caption{Detailed Hits@3 results on the Q2B datasets testing generalization and deductive reasoning.}
\label{table_9}
\end{table*}

\end{document}